\documentclass[lettersize,journal]{IEEEtran}
\usepackage{amsmath,amsfonts}
\usepackage{algorithmic}
\usepackage{algorithm}
\usepackage{array}
\usepackage{textcomp}
\usepackage{amsmath}
\usepackage{stfloats}
\usepackage{url}
\usepackage{verbatim}
\usepackage{graphicx}
\usepackage{cite}
\usepackage{multirow}
\usepackage{subfigure}
\usepackage{booktabs}
\usepackage{color}

\begin{document}

% \title{ENDAE: Edge and Node Aware Dual AutoEncoder for Traffic Event Detection}
\title{Intention-aware Denoising Diffusion Model for Trajectory Prediction} 
\author{Chen Liu, Shibo He, Haoyu Liu, Jiming Chen
\thanks{Chen Liu, Shibo He, Jiming Chen are with State Key Lab. of Industrial Control Technology, Zhejiang University, Hangzhou 310027, China. Email: \{liu777ch, s18he, cjm\}@zju.edu.cn.
Haoyu Liu is with Fuxi AI Lab, NetEase Games, Hangzhou 310052, and also with State Key Laboratory of Industrial Control Technology, Zhejiang University, Hangzhou 310027, China. Email: liuhaoyu03@corp.netease.com.
}}
% \title{Intention May Provide a Shortcut for Diffusion Model in Trajectory Prediction}
% \title{Explicitly Modelling Edge-level Event Makes Better Traffic Event Detector}
% \author{Chen Liu, Jiming Chen, Haoyu Liu, Shizhong Li, Shibo He
% \thanks{Chen Liu, Jiming Chen, Shizhong Li, Shibo He are with State Key Lab. of Industrial Control Technology, Zhejiang University, Hangzhou 310027, China. Email: \{lc\_nj, cjm, lisz, s18he\}@zju.edu.cn.

% Haoyu Liu is with Fuxi AI Lab, NetEase Games, Hangzhou 310052, and also with State Key Laboratory of Industrial Control Technology, Zhejiang University, Hangzhou 310027, China. Email: liuhaoyu03@corp.netease.com.
% }}
% \thanks{Chen Liu, Jiming Chen, Shibo He are with State Key Lab. of Industrial Control Technology, Zhejiang University, Hangzhou 310027, China. Email: \{lc\_nj, cjm, s18he\}@zju.edu.cn.

% H. Liu is with the State Key Laboratory of Industrial Control Technology, Zhejiang University, Hangzhou 310027, China, and also with Fuxi AI Lab, NetEase Games, Hangzhou 310052, China (e-mail: haoyu\_liu@zju.edu.cn).

% H. Dong is with the State Key Laboratory of Rail Traffic
% Control and Safety, Beijing Jiaotong University, Beijing 100044, China
% (e-mail: hrdong@bjtu.edu.cn).}
% % \thanks{This paper was produced by the IEEE Publication Technology Group. They are in Piscataway, NJ.}% <-this % stops a space
% % \thanks{Manuscript received April 19, 2021; revised August 16, 2021.}}

\markboth{Journal of \LaTeX\ Class Files,~Vol.~14, No.~8, August~2021}%
{Shell \MakeLowercase{\textit{et al.}}:Hierarchical Diffusion Model}


\maketitle

\begin{abstract}
Trajectory prediction is an essential component in autonomous driving, particularly for collision avoidance systems. Considering the inherent uncertainty of the task, numerous studies have utilized generative models to produce multiple plausible future trajectories for each agent. However, most of them suffer from restricted representation ability or unstable training issues. To overcome these limitations, we propose utilizing the diffusion model to generate the distribution of future trajectories. Two cruxes are to be settled to realize such an idea. 
First, the diversity of intention is intertwined with the uncertain surroundings, making the true distribution hard to parameterize. Second, the diffusion process is time-consuming during the inference phase, rendering it unrealistic to implement in a real-time driving system. 
We propose an \underline{I}ntention-aware denoising \underline{D}iffusion \underline{M}odel (IDM), which tackles the above two problems. We decouple the original uncertainty into intention uncertainty and action uncertainty, and model them with two dependent diffusion processes. To decrease the inference time, we reduce the variable dimensions in the intention-aware diffusion process and restrict the initial distribution of the action-aware diffusion process, which leads to fewer diffusion steps. 
To validate our approach, we conduct experiments on Stanford Drone Dataset (SDD) and ETH/UCY dataset. Our methods achieve state-of-the-art results, with an FDE of 13.83 pixels on SDD dataset and 0.36 meters on ETH/UCY datasets. Compared with the original diffusion model, IDM reduces inference time by two-thirds. 
Interestingly, our experiments further reveal that introducing intention information is beneficial in modeling the diffusion process of fewer steps.

\end{abstract}

\begin{IEEEkeywords}
trajectory prediction, diffusion models
\end{IEEEkeywords}

\section{Introduction} \label{intro}
% traffic anomaly detection很重要
\IEEEPARstart{A}{lthough} high-level autonomous driving systems can significantly enhance our convenience, it remains a challenging task to maintain its security \cite{huang2022survey}. 
% To overcome this challenge, the perception subsystem helps the vehicle understand its surroundings, while the planning subsystem creates a safe and appropriate maneuver for the vehicle \cite{wen2023tofg}. Trajectory prediction serves as a crucial component that links the two subsystems. 
As mentioned in \cite{zhu2023reciprocal,wang2022probabilistic,rudenko2020human}, self-driving vehicles must
anticipate the future movements of surrounding agents (including vehicles and pedestrians) in order to plan proactive motions and avoid collision with them. 
Therefore, trajectory prediction plays a crucial role in autonomous driving systems \cite{hu2022trajectory}.

% Early works in trajectory prediction predominantly rely on deterministic methods, which offer a single predicted trajectory for each agent \cite{helbing1995social,xue2018ss,wang2020improving,gao2020vectornet}. 
Traditional deterministic trajectory prediction methods aim to provide a single future trajectory for each agent \cite{helbing1995social,xue2018ss,wang2020improving,gao2020vectornet}. 
However, it has been recognized that not all relevant clues, such as the intentions and habits of the agents, can be fully acquired. Consequently, there may exist multiple plausible future trajectories for an agent \cite{zhao2021tnt}, which can be ignored by these deterministic methods. 
% As a result, traditional deterministic trajectory prediction \cite{helbing1995social,xue2018ss,wang2020improving,gao2020vectornet} only providing a single prediction for each agent fail to encompass all possible paths of future trajectories \cite{trentin2023multi}. 
To address this issue, probabilistic trajectory prediction attempts to generate multiple predictions to encompass all possible
outcomes, and it has garnered growing interest in recent years \cite{trentin2023multi}.

% On the one hand, the interaction between the target agent and its complex surroundings must be taken into consideration, such as pedestrians, vehicles, and infrastructures. 
Previous studies have investigated different generative models for probabilistic trajectory prediction, including noise-based models, bivariate-Gaussian (BG)-based models, conditional variational autoencoder (CVAE)-based models, and generative adversarial network (GAN)-based models \cite{huang2023multimodal}.
Noise-based models, used by Gupta et~al., Thiede et~al., and Deo et~al. \cite{gupta2018social,thiede2019analyzing,deo2022multimodal} involve injecting random noise into neural networks to generate multiple future trajectories and training the networks using a variety loss. However, the variety loss fails to penalize unrealistic trajectories, resulting in the inability to accurately describe the true distribution of possible trajectories \cite{guo2022end}. 
BG-based models, used by Alal et~al. \cite{alahi2016social} and Mohamd et~al. \cite{mohamed2020social} assume that future positions follow a bivariate-Gaussian distribution and estimate the distribution parameters using maximum likelihood estimation. However, real-world trajectories may follow more complex distributions, leading to the capped performance of their methods \cite{huang2023multimodal}. 
% The grid-based model divides the scene into grid cells and predicts the occupancy probability of each cell in the future. Although this method achieves outstanding performance, it requires significant effort to construct and train the trajectory map \cite{mangalam2021goals}
CVAE-based models aim to incorporate uncertainty into the predictions by introducing a latent distribution \cite{salzmann2020trajectron++,ben2022raising,liu2021social,chen2021personalized}. Nevertheless, CVAE is observed to generate unnatural trajectories due to its limited ability to model complex distributions \cite{mangalam2021goals}.  
Additionally, GANs are employed to fit parameterized probability distributions \cite{dendorfer2021mg,fang2020tpnet,kosaraju2019social}. However, the unstable training procedure of GANs hinders their practical development \cite{liang2022stglow}. 
In summary, all these generative models suffer from limited ability to represent sophisticated trajectory distributions or unstable training processes.

\begin{figure}[t]
\centering
\subfigure[Intention uncertainty]{
\includegraphics[width=4.3cm]{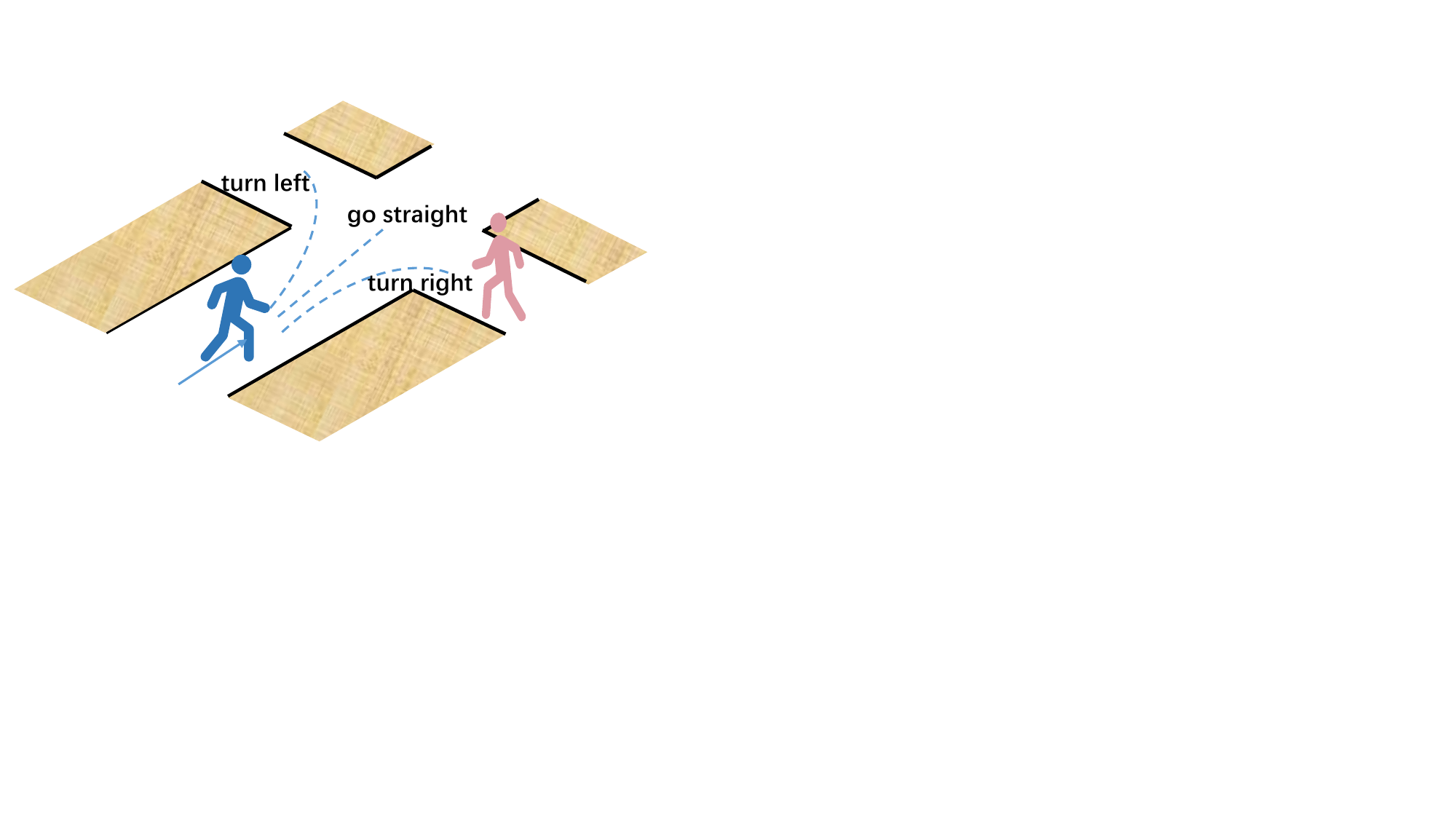}
%\caption{fig1}
\label{fig:mot1_a}
}%
\subfigure[Action uncertainty]{
\includegraphics[width=4.3cm]{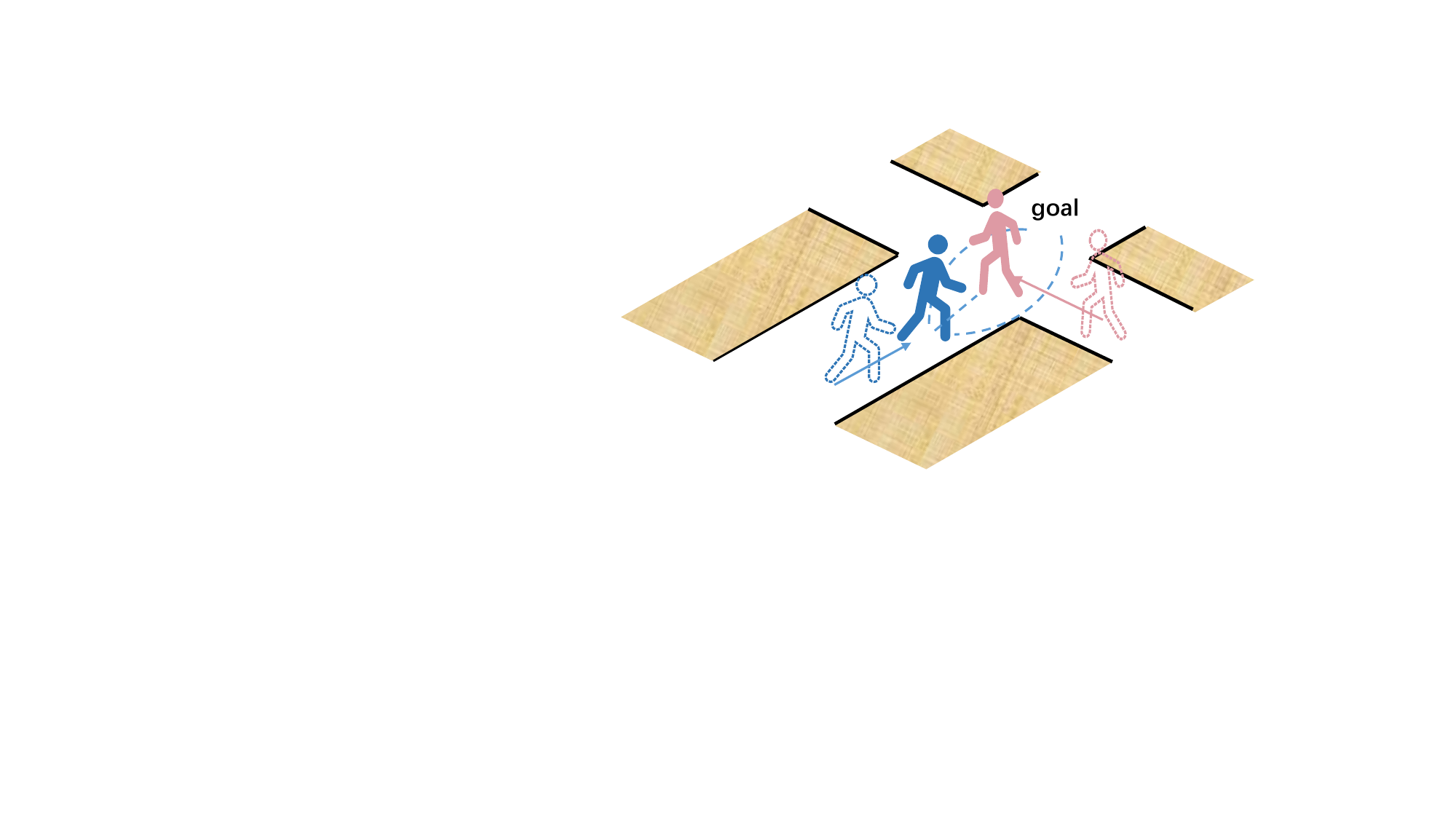}
%\caption{fig2}
\label{fig:mot1_b}
}%
\centering
\caption{Two kinds of uncertainty: (a) A pedestrian may turn left, turn right, or proceed straight based on his own will; (b) A pedestrian with a deterministic goal can also choose different paths in order to avoid collisions with their surroundings.}
\label{fig:mot1}
\end{figure}

Recently, diffusion models have shown their potential in overcoming two aforementioned limitations and have gained prominence among generative models  \cite{choi2021ilvr,dhariwal2021diffusion,nichol2021improved}. 
% in particular, thanks to their more powerful representation ability and stable training process \cite{ho2016generative}.
Compared with CVAE and BG-based models, diffusion models offer greater representation ability, as they can generate complex distributions through a series of steps \cite{ho2016generative}. Additionally, the training process of diffusion models is more stable compared to GANs \cite{gu2022stochastic}.
As a result, diffusion models have achieved state-of-the-art performance in various domains, including image synthesis\cite{choi2021ilvr,dhariwal2021diffusion,nichol2021improved}, video generation\cite{ho2022video,harvey2022flexible} and audio processing\cite{chen2020wavegrad,kong2020diffwave}. Considering the aforementioned advantages of diffusion models, we propose adopting diffusion models for the trajectory prediction problem.
% In the diffusion model, future positions of agents are treated as particles in thermodynamics, with the trajectory viewed as the initial state of particles. During the diffusion process, the trajectory spreads until it encompasses all possible regions. During the inference stage, we reverse this diffusion process to generate predicted trajectories from stochastic positions. 
However, two main challenges need to be addressed to realize this idea.

% \begin{figure}[t]
% \centering

% \includegraphics[width=6.6cm]{pictures/mot0.pdf}
% %\caption{fig2}

% \caption{Performance on the traffic events happened within and outside the sensor coverage. All the SOTAs perform worse on the events outside the coverage.}
% \label{fig:mot0}
% \end{figure}

% Inspired by the above issues, we propose to improve the detection performance via a more fine-grained modeling strategy which includes sensing range as an important hyper-parameter. However, there exist multiple challenges to realize such an idea:

\textbf{{\em How to clearly model the uncertainty of multimodal future?}} 
% The largest challenge of the trajectory prediction task lies in the uncertainty of human motion or driver's manipulation, which is interweaved by internal stimulus and external environment factors. 
% One of the main challenges of trajectory prediction lies in how to accurately model the uncertainty of multi-modal future trajectories. 
The uncertainty of agent movement is significantly influenced by both internal stimulus and external environmental factors.
For example, at a crossroad, individuals may choose to turn left, turn right, or proceed straight based on their intentions, as depicted in Fig.~\ref{fig:mot1_a}. Furthermore, even when the goal (endpoint) is deterministic, they may select different paths. Fig.~\ref{fig:mot1_b} illustrates a scenario where an individual intends to go straight but adjusts their movement to the left or right to avoid a collision with another person.   
It has been recognized in \cite{zhao2021tnt} that the future trajectory distribution should be multimodal, with each mode representing a different intention of the agent. 
Applying the original diffusion model to this task would neglect the multimodal property of the trajectory \cite{gu2022stochastic}, as it only models the uncertainty of paths without considering intention information. As a result, it may produce stochastic and unnatural predictions.
% However, it is difficult for a diffusion model which denoises from a Gaussian distribution to learn a multi-modal distribution. 
% In our work, we address this challenge by decoupling the uncertainty into intention uncertainty and action uncertainty.
% To model the intention uncertainty, we utilize a diffusion process to estimate the distribution of the agent’s goal. To model the action uncertainty, we employ another diffusion process to estimate the distribution of paths conditioned on a specific goal. 
% % We then utilize two dependent diffusion processes to model these two kinds of uncertainty, respectively.
% During the prediction stage, we first infer multiple possible goals of the agent and then generate trajectories conditioned on all of these inferred goals. This approach allows for capturing the multimodal nature of the trajectory distribution, which can both integrate intention and action uncertainties.

% \textbf{{\em How to model the continuous and dynamic intention without anchors?}}

% \textbf{{\em How to make the predictions aware of the intention?}}
% Previously, the mode variable are merged with the total encodings for the target vehicle, and are fed into an MLP-based decoder, which brings unnatural trajectories. In this paper, we propose to generate the prior trajectory distribution conditioned on a specific goal. Then the surrounding information will add perturbation to the prior trajectory. A gated prior network is devised to produce both linear and non-linear trajectory given a goal. 

\textbf{{\em How to enhance the efficiency of diffusion model?}}
The diffusion model typically performs poorly in terms of time efficiency, resulting in a significant drawback when applied to trajectory prediction for autonomous driving systems. The considerable time consumption can be attributed to two factors. Firstly, the diffusion model follows a Markov chain, where the ground truth distribution is transformed into a prior Gaussian noise distribution by injecting noises step by step. To ensure that the prior noise distribution follows a Gaussian distribution, the diffusion process entails a large number of steps ~\cite{meng2021sdedit}. Consequently, during the reverse process, obtaining the ground truth distribution from the Gaussian noise also necessitates a large number of denoising steps, resulting in low efficiency. Secondly, the inefficiency is further exacerbated by the high dimensions of data involved in the diffusion process \cite{rombach2022high}. Dealing with high-dimensional data requires more computational resources and time, making the overall process slower.
% Therefore, in our work, we make two efforts to shorten the inference time of the diffusion model.
% Firstly, we reduce the dimension of the variable being modeled. We consider each trajectory’s endpoint as the particle instead of the entire trajectory. Secondly, instead of assuming a normal prior distribution,  we devise a PriorNet to estimate the specific prior distribution of the diffusion process. This approach allows us to reduce the number of diffusion steps. In addition, we design a corresponding loss function to facilitate training of the entire process in an end-to-end manner. We also discover the importance of intention in providing valuable clues for estimating the prior distribution more accurately.

% our works
To address the aforementioned challenges, we propose an intention-aware diffusion model called IDM for trajectory prediction. 
Our approach involves decoupling the uncertainty of trajectory prediction into intention uncertainty and action uncertainty. To model the intention uncertainty, we utilize a diffusion process to estimate the distribution of the agent’s goal. For action uncertainty, we employ another diffusion process to estimate the distribution of paths conditioned on a specific goal. During the prediction stage, we first infer multiple possible goals of the agent and then generate trajectories conditioned on all of these inferred goals. This approach allows for capturing the multimodal nature of the trajectory distribution, which can integrate both intention and action uncertainties. 
% The model consists of two diffusion processes, where the first one generates the agent's goals while the second one generates the goal-conditioned trajectories. 
% Specifically, We devise a gated prior network to produce ideal trajectory conditioned on current position and goal location.
To enhance computational efficiency, in the goal diffusion process, we consider each trajectory’s endpoint as the particle instead of the entire trajectory, thus reducing the dimension of the data being modeled. In the path diffusion process, rather than using prior Gaussian noise distribution, we propose to estimate the prior noise distribution by a neural network, thereby reducing the number of steps required in the diffusion process. Additionally, we design a corresponding loss function to enable end-to-end training of the two diffusion processes. We evaluate our model on two real-world datasets, and the results demonstrate IDM’s superior performance. It achieves state-of-the-art results in both Average Displacement Error (ADE) and Final Displacement Error (FDE). Compared to traditional denoising diffusion models, our model reduces inference times by two-thirds. Interestingly, we also discover the significance of intention in providing valuable clues for estimating the prior noise distribution accurately. The contributions are summarized as follows:

% In general, our contributions are listed as follows:
\begin{itemize}
\item We propose a two-stage diffusion model for trajectory generation called IDM, which decouples the uncertainty of trajectory prediction into goal uncertainty and action uncertainty, and model them by two dependent diffusion process.

\item To enhance computational efficiency, we propose a PriorNet for estimating prior noise distribution and utilize a tailored loss function during training, which significantly reduces the required number of steps in the diffusion process. Interestingly, we also find that the goal can provide valuable information for accurate prior distribution noise estimation.

\item We perform extensive experiments on two real-world datasets. Our findings show that IDM achieves state-of-the-art detection performance in terms of ADE and FDE. Additionally, IDM significantly reduces inference time by about two-thirds compared to the original diffusion model.

\end{itemize}

The rest of this paper is organized as follows. In Section II, we provide a comprehensive review of related works on trajectory prediction and diffusion models. Section III formulates the problem and provides preliminary knowledge. Section IV presents a detailed explanation of our proposed method. In Section V, we conduct experiments to assess the performance of our method. Finally, we conclude our study in Section VI.

\section{Related Work}

\subsection{Trajectory Prediction}

Trajectory prediction plays a crucial role in the field of self-driving vehicles. Previous studies in this area can be broadly categorized into three main approaches: physics-based, planning-based, and pattern-based \cite{rudenko2020human}.

\subsubsection{Physics-based approaches} 
Previous works on trajectory prediction use hand-crafted dynamic models based on Newton’s laws of motion.
Early approaches often employ methods like autoregressive models \cite{elnagar1998motion}, Kalman filters\cite{barth2008will}, and particle filters\cite{cai2006robust} to make one-step-ahead predictions. These predictions are based on classical kinematic models like the constant velocity model (CV), the constant acceleration model (CA), the bicycle model and so on \cite{elnagar2001prediction,schubert2008comparison,mogelmose2015trajectory}. Other works incorporate map-related contextual clues\cite{yang2005nonlinear,batkovic2018computationally,petrich2013map} and social interaction information\cite{yan2014modeling,karamouzas2009predictive,zanlungo2011social} into physics-based models. While these methods perform well under mild conditions, their performance deteriorates over longer prediction horizons due to the absence of future control signals, limiting their applicability to long-term prediction tasks of more than 1 second \cite{huang2022survey}.

\subsubsection{Planning-based approaches}
Planning-based approaches assume that agents are rational decision-makers during action. Therefore, they turn the prediction problem into a sequential decision-making problem aimed at finding an optimal motion sequence that minimizes a cost function \cite{rudenko2020human}. 
The approaches can be classified into forward-planning ones and inverse-planning ones. 
The forward-planning methods generate plausible trajectories using path planning with hand-crafted cost-function \cite{gong2011multi}. The interactions among multi-agents are also considered through cooperative planning in joint state-space \cite{bahram2015game,chen2017decentralized,rosmann2015timed}. However, these works all predefine an explicit cost or reward function, which sometimes obviates from reality. The inverse planning methods estimate the cost function from all observations via Inverse Reinforcement Learning (IRL) \cite{previtali2016predicting}.
Other works employ Generative Adversarial Imitation Learning (GAIL) \cite{ho2016generative} to model the future motion state distribution directly without learning the reward function first \cite{kuefler2017imitating,li2017infogail}. These methods are computationally intensive and consume large training costs.

\subsubsection{Pattern-based approaches} 
With the development of deep learning, researchers develop approaches that learn trajectory patterns directly from data. Most of the existing deep learning models for trajectory prediction adopt an encoder-decoder architecture \cite{sutskever2014sequence}. The encoder captures meaningful information for prediction, such as historical trajectories, map information, and states of neighbor agents. To model historical trajectories, previous works utilize sequence modeling techniques like long short-term memory networks (LSTM) \cite{alahi2016social,zhang2023spatial}, temporal convolutional network (TCN) \cite{li2021spatial}, and Transformer\cite{deo2022multimodal}. To provide prior knowledge of road structure, the high-definition map can be encoded by rasterization\cite{chai2019multipath} or vectorization \cite{gao2020vectornet}. The interactions among multiple agents are often modeled by the social pooling mechanism \cite{bae2022learning}, attention mechanisms \cite{messaoud2020attention,zhang2022ai,li2023interaction} or graph neural networks \cite{xu2022groupnet,shi2021sgcn,mohamed2020social}. The decoder network aggregates all information and generates the eventual predictions. Song et~al. focus on the feasibility of predictions, by imposing dynamic and scene-compliant restrictions on the output sequence \cite{song2022learning,chen2022scept,phan2020covernet}. Other works pay attention to stochastic predictions. A common practice is to incorporate a noise signal into the decoder and utilize the variety loss \cite{gupta2018social} to train the decoder. However, Guo et~al. point out that the variety loss can not penalize those unrealistic trajectories, and may generate unfeasible samples during inference \cite{guo2022end}. Therefore, the following works turn to adopt generative models. For example, Salzmann et~al. firstly propose a Conditional Variational Autoencoder (CVAE) based method Trajectron++ \cite{salzmann2020trajectron++}, and several works
follow the architecture \cite{ben2022raising,liu2021social,chen2021personalized}. Other works apply Generative Adversarial Network (GAN) to trajectory generation \cite{dendorfer2021mg,fang2020tpnet,kosaraju2019social}. To generate multimodal trajectories, the latent variable which implies the intention of the agents such as goal endpoints\cite{chiara2022goal,mangalam2021goals} and goal lanes \cite{deo2022multimodal,wang2022ltp} are estimated to enhance the training \cite{zhao2021tnt,gu2021densetnt}. However, the CVAE-based methods encounter the performance bottleneck when handling complicated distribution, while the GAN-based methods are trapped in unstable training. In this work, we aim to employ diffusion models to achieve multimodal and stochastic prediction. The most related works to us are \cite{gu2022stochastic} and \cite{mao2023leapfrog}. Our work differs from them in two aspects. Firstly, we consider the multimodal property and devise a goal-aware diffusion process. Secondly, we significantly reduce the number of steps in the diffusion process by estimating the multi-modal prior distribution instead of using the normal distribution assumption, which significantly shortens the inference time. 
% Although Mao et~al.~\cite{mao2023leapfrog} also adopt fewer steps in the diffusion process, they have two limitations. Firstly, they employ a two-stage training strategy where the diffusion model with large steps should be trained first, which is time-consuming. Secondly, they assume that the initial distribution still follows a Gaussian distribution, which may lead to poor performance. In this paper, we directly train a diffusion model with fewer steps in an efficient manner. Additionally, instead of using the aforementioned Gaussian distribution assumption, we find that initializing the noise distribution with a multi-modal distribution benefits the prediction. The details are introduced in Sec.~\ref{ssec:tdp}.
% Thirdly, our model can be applied to any architecture that estimates the agent’s intention.

\subsection{Diffusion Model}
Recent years have witnessed the success of diffusion model in various domains such as image synthesis\cite{choi2021ilvr,dhariwal2021diffusion,nichol2021improved}, video generation\cite{ho2022video,harvey2022flexible}, sequence modeling \cite{li2022diffusion,tashiro2021csdi} and audio processing\cite{chen2020wavegrad,kong2020diffwave}. As a powerful and novel deep generative model, the diffusion model originates from non-equilibrium thermodynamics and is first proposed in \cite{sohl2015deep}. The key concept of the diffusion model is to turn original signals into a known noise distribution by injecting noise gradually and then reverse the process during generation \cite{yang2022diffusion}. The denoising process is modeled by a parameterized Markov chain, which can be learned by a neural network. Compared with classical generative models like VAE\cite{rezende2014stochastic} and GAN\cite{goodfellow2020generative}, the diffusion model shows great potential in representation learning and has a stable training procedure with solid theoretical supports \cite{cao2022survey}. However, the diffusion model has an inherent weakness of slow generation process \cite{luo2022understanding}, which limits its application in real-time scenarios such as autonomous driving. 
To address this issue, we propose an efficient diffusion model for trajectory prediction, taking inspiration from previous works. Robin et al. propose a diffusion process targeting low-dimension latent features to reduce the cost of training and inference \cite{rombach2022high}. Therefore, utilizing intention (goal) instead of the whole trajectory to establish the diffusion process can result in significant cost savings. Furthermore, Meng et al. claim that the diffusion process removes signals from high-frequency to low-frequency \cite{meng2021sdedit}. 
% thus reducing the number of steps needed to generate samples by utilizing low-frequency information in the noise distribution . 
In our proposed framework, we assume that the agent’s intention (goal) contains low-frequency information, while the path contains high-frequency information. Therefore, we utilize the predicted goal information to estimate the intermediate distribution with low-frequency signals. This allows us to model the process from the original distribution to this intermediate distribution, effectively reducing the number of steps required in the diffusion process.
% Our work focuses on finding a low-dimension latent variable containing low-frequency information in trajectory prediction tasks. Considering that the agent’s intention may meet the requirements, we design an intention-aware diffusion model to enhance efficiency and accuracy of trajectory prediction.

% In this work, our strategy falls into the same direction of replay-based methods.

% \begin{figure}[t]
% 	\centering
% 	% Requires \usepackage{graphicx}
% 	\includegraphics[width=0.5\textwidth]{pictures/problem.pdf}\\
% 	\caption{}
% 	\label{fig:problem}
% \end{figure}

\section{System Model} \label{sec:pre}

\subsection{Problem Statement} \label{sec:pro}
The goal of trajectory prediction is to predict the future locations of agents on the road, such as pedestrians, riders, or vehicles, based on their historical tracepoints. Due to the inherent uncertainty of moving objects, there are many plausible routes that the agents could follow in the future, so our focus is on generating a trajectory distribution that is as realistic as possible, guided by all available observations.
Mathematically, given a target agent with trajectories in the past $x=\left\{s_{t}\in \mathbb{R}^{2} \mid t=-T_{P}+1, -T_{P}+2,...,0 \right\}$, where $s_t$ denotes its 2-D locations at time $t$, and $T_{P}$ is the time step of trajectory, we aim to predict its future trajectory distribution $y=\left\{ p(s_t)\mid t=1,2...,T_{Q} \right\}$, where $T_{Q}$ is the length of prediction, and $p(s_t)$ represents the distribution of all possible locations at time $t$. In realistic scenario, we could predict $\mathcal{K}$ trajectories $\left\{y_{i} \in \mathbb{R}^{T_{Q}\times2} \mid i=1,2,...\mathcal{K}\right\}$ in the future. In addition, environmental factors such as historical locations of neighbor agents $e=\left\{s_t^{ne}\mid t=-T_{P}+1, -T_{P}+2,...,0 \right\}$, and the map $\mathcal{M}$ can be vectorized \cite{gao2020vectornet} to assist the prediction. We summarize all the notations used in our work in Table~\ref{tab:nota}.

\begin{table}[h]
\centering
\caption{Summary of Notations}
\begin{tabular}{ll}
\hline
Symbol & Description       \\ \hline
\multicolumn{1}{c}{$x$}      & The agent's historical trajectory  \\
\multicolumn{1}{c}{$y$}      & The agent's future trajectory     \\
\multicolumn{1}{c}{$e$}      & The neighbor's historical trajectory\\ 
\multicolumn{1}{c}{$\mathcal{M}$}  & The map of the scene \\
\multicolumn{1}{c}{$T_P$}      & Number of historical steps  \\
\multicolumn{1}{c}{$T_Q$}      & Number of future steps  \\
\multicolumn{1}{c}{$\mathcal{K}$}  & Number of future predictions \\
\multicolumn{1}{c}{$K$}  & Number of steps in the goal diffusion process \\
\multicolumn{1}{c}{$S$}  & Number of steps in the trajectory diffusion process \\ 
\multicolumn{1}{c}{$\alpha_{k}^{g}$}  & The diffusion coefficient at $k$-th step for goal \\
\multicolumn{1}{c}{$\alpha_{s}^{t}$}  & The diffusion coefficient at $s$-th step for trajectory \\
\multicolumn{1}{c}{$g_\psi$}  & Encoder with parameters $\psi$ \\
\multicolumn{1}{c}{$\epsilon_\theta$}  & EndNet with parameters $\theta$ \\
\multicolumn{1}{c}{$\mu_\phi$}  & PriorNet with parameters $\phi$ \\
\multicolumn{1}{c}{$\epsilon_\varphi$}  & PathNet with parameters $\varphi$ \\
\multicolumn{1}{c}{$\lambda_1$}  & The coefficient for trajectory diffusion loss \\
\multicolumn{1}{c}{$\lambda_2$}  & The coefficient for trajectory prior loss \\
\multicolumn{1}{c}{$\mathbf{x}$}  & The context vectors generated by the encoder \\
\multicolumn{1}{c}{$\mathbf{c}_{k}$}  & The goal at $k$-th step in diffusion process \\
\multicolumn{1}{c}{$\mathbf{y}_{s}$}  & The trajectory at $s$-th step in diffusion process \\
\multicolumn{1}{c}{$\mathbf{c}_{0}$}  & The ground truth goal \\
\multicolumn{1}{c}{$\mathbf{y}_{0}$}  & The ground truth trajectory \\
\hline
\end{tabular}
\label{tab:nota}
\end{table}

\begin{figure}[t]
\centering
\includegraphics[width=0.4\textwidth]{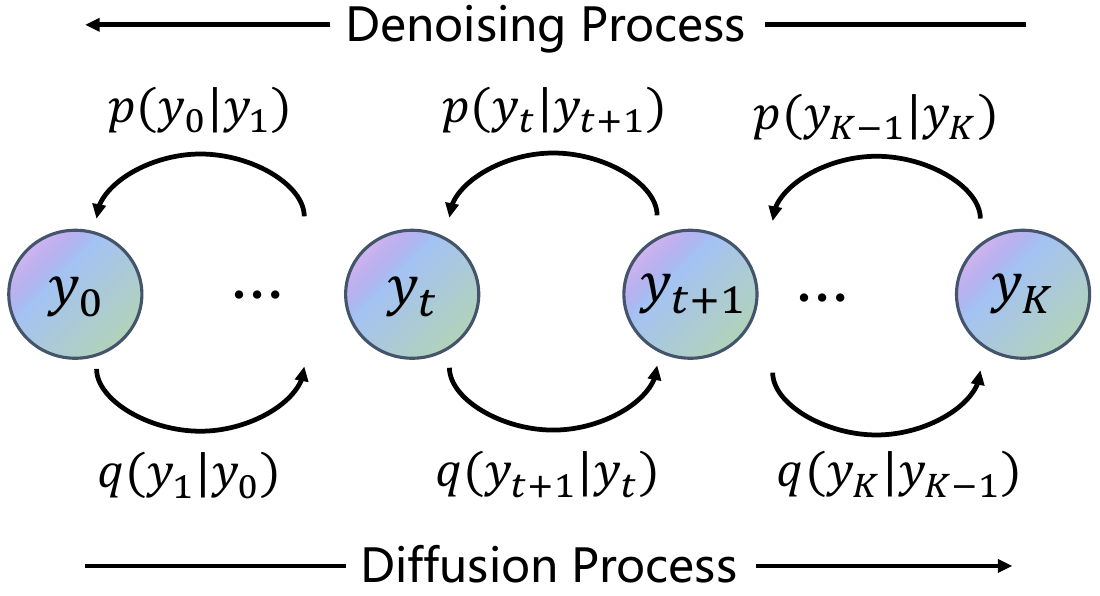}\\
\caption{Denoising Diffusion Probabilistic Model}
\label{fig:ddpm}
\end{figure}

\subsection{Denoising Diffusion Probabilistic Model}
Generative models aim to discover the real distribution of observed samples by representing uncertainty through latent variables. 
Inspired by non-equilibrium thermodynamics, the denoising diffusion probabilistic model (DDPM) describes the distribution of samples as particles in motion, akin to the behavior of particles in thermodynamics. As noise is gradually added to the observed samples, the distribution of real data, which initially has low uncertainty, transforms into a random distribution with high uncertainty. This transformation process is commonly referred to as the diffusion process or forward process. As shown in Fig~\ref{fig:ddpm}, the diffusion process is modeled using a parameterized Markov chain:
\begin{equation}
\begin{aligned}
q\left(\mathbf{y}_{1: K} \mid \mathbf{y}_0\right)
&:=\prod_{k=1}^K q\left(\mathbf{y}_k \mid \mathbf{y}_{k-1}\right) \\
q\left(\mathbf{y}_k \mid \mathbf{y}_{k-1}\right)
&:=\mathcal{N}\left(\mathbf{y}_k ; \sqrt{1-\beta_t} \mathbf{y}_{k-1}, \beta_k \mathbf{I}\right)
\end{aligned}
\end{equation}
where $\beta_1,\beta_2,...\beta_k$ denote the variance schedulers that control the level of Gaussian noise that is injected into the original signals. 
On the other hand, the reverse process, also known as the denoising process, involves transforming the noisy distribution back into the sample distribution. This process can also be represented as a Markov chain:
\begin{equation}
\label{eq2}
\begin{aligned}
p_\theta\left(\mathbf{y}_{0: K}\right)&:=p\left(\mathbf{y}_K\right) \prod_{k=1}^K 
p_\theta\left(\mathbf{y}_{k-1} \mid \mathbf{y}_k\right) \\
\quad p_\theta\left(\mathbf{y}_{k-1} \mid \mathbf{y}_k\right)&:=\mathcal{N}\left(\mathbf{y}_{k-1} ; \boldsymbol{\mu}_\theta\left(\mathbf{y}_k, k\right), \boldsymbol{\Sigma}_\theta\left(\mathbf{y}_k, k\right)\right)
\end{aligned}
\end{equation}
where the mean $\boldsymbol{\mu}_\theta\left(\mathbf{x}_k, k\right)$ and variance $\boldsymbol{\Sigma}_\theta\left(\mathbf{x}_k, k\right)$ of reverse transition is learnt by a neural network. 

Based on the fixed forward process parameters, We can acquire $y_t$
at any time step as follows: 
\begin{equation}\label{eq3}
q\left(\mathbf{y}_k \mid \mathbf{y}_0\right)=\mathcal{N}\left(\mathbf{y}_k ; \sqrt{\bar{\alpha}_k} \mathbf{y}_0,\left(1-\bar{\alpha}_k\right) \mathbf{I}\right)
\end{equation}
using the notations: $ \alpha_k:=1-\beta_k $ and $\bar{\alpha}_k:=\prod_{s=1}^k \alpha_s$. It can be seen that the final noised distribution $y_k$ approaches the normal distribution as $k$ increases. A common practice is to set a large step number $K$, which can result in $y_K$ being equivalent to a normal distribution.  By using this setting, we can obtain real samples from the normal distribution through the reverse process.

Our aim is to learn $p_\theta\left(\mathbf{y}_{k-1} \mid \mathbf{y}_k\right)$ that match the diffusion process. The training objective is to maximize the likelihood function $\mathbb{E}\left[\log p_\theta\left(\mathbf{y}_0\right)\right]$, using variational inference: 
\begin{equation}
\begin{aligned}
\mathbb{E}\left[\log p_\theta\left(\mathbf{y}_0\right)\right] 
& = \mathbb{E}_{q\left(\mathbf{y}_{1:K} \mid \mathbf{y}_0\right)}\left[\log \frac{p_\theta\left(\mathbf{y}_{0: K}\right)}{p_\theta\left(\mathbf{y}_{1: K} \mid \mathbf{y}_0\right)}\right] \\
& = \mathbb{E}_{q\left(\mathbf{y}_{1:K} \mid \mathbf{y}_0\right)}\left[\log \frac{p_\theta\left(\mathbf{y}_{0: K}\right)}{q\left(\mathbf{y}_{1: K} \mid \mathbf{y}_0\right)}\right] \\
& + D_{K L}\left(q\left(\mathbf{y}_{1:K} \mid \mathbf{y}_0,\right)\| p_\theta\left(\mathbf{y}_{1: K} \mid \mathbf{y}_0\right)\right)
\end{aligned}
\end{equation}
then we can maximize the first term called evidence lower bound:
\begin{equation}
% \mathbb{E}_q\left[\log \frac{p_\theta\left(\mathbf{y}_{0: K}\right)}{q\left(\mathbf{y}_{1: K} \mid \mathbf{y}_0\right)}\right] \\
ELBO
=\mathbb{E}_q\left[\log p\left(\mathbf{y}_K\right)+\sum_{k=1}^K \log \frac{p_\theta\left(\mathbf{y}_{k-1} \mid \mathbf{y}_k\right)}{q\left(\mathbf{y}_k \mid \mathbf{y}_{k-1}\right)}\right]
\end{equation}
since $y_K$ is determined by the real data distribution $y_0$ and the predefined variance schedulers, it represents a constant distribution and can be disregarded during training. Therefore, the loss function is defined as follows:
\begin{equation}
\begin{aligned}
L(\theta) 
& =\mathbb{E}_q\left[\sum_{k=1}^K \log \frac{p_\theta\left(\mathbf{y}_{k-1} \mid \mathbf{y}_k, \right)}{q\left(\mathbf{y}_k \mid \mathbf{y}_{k-1}\right)}\right] \\
& =\mathbb{E}_q\left[\sum_{k=1}^K D_{K L}\left(q\left(\mathbf{y}_{k-1} \mid \mathbf{y}_k, \mathbf{y}_0\right) \| p_\theta\left(\mathbf{y}_{k-1} \mid \mathbf{y}_k\right)\right)\right]
\end{aligned}
\end{equation}
Indeed, posterior distribution $q\left(\mathbf{y}_{k-1} \mid \mathbf{y}_k, \mathbf{y}_0\right)$ is also a Gaussian distribution:
\begin{equation}
\begin{aligned}
q\left(\mathbf{y}_{k-1} \mid \mathbf{y}_k, \mathbf{y}_0\right)&=\mathcal{N}\left(\mathbf{y}_{k-1} ; \tilde{\boldsymbol{\mu}}_k\left(\mathbf{y}_k, \mathbf{y}_0\right), \tilde{\beta}_k \mathbf{I}\right)
\end{aligned}
\end{equation}
where the parameters are calculated according to:
\begin{equation}
\begin{aligned}
\tilde{\boldsymbol{\mu}}_k\left(\mathbf{y}_k, \mathbf{y}_0\right) & =\frac{\sqrt{\bar{\alpha}_{k-1}} \beta_k}{1-\overline{\alpha_k}} \mathbf{y}_0+\frac{\sqrt{\alpha_k}\left(1-\bar{\alpha}_{k-1}\right)}{1-\bar{\alpha}_k} \mathbf{y}_k \\
&=\frac{1}{\sqrt{\alpha_k}}\left(\mathbf{y}_k-\frac{\beta_k}{\sqrt{1-\bar{\alpha}_k}} \epsilon\right) \\ 
\tilde{\beta}_k & =\frac{1-\bar{\alpha}_{k-1}}{1-\bar{\alpha}_k} \beta_k \mathbf{I} .
\end{aligned} 
\end{equation}
where $ \epsilon \sim\mathcal{N}\left(0,\mathbf{I}\right)$.  Based on (\ref{eq2}), the loss function can be regarded as the KL divergence of two Gaussian distributions. To simplify the optimization, we set $\boldsymbol{\Sigma}_\theta\left(\mathbf{y}_k, k\right)=\tilde{\beta}_k$, and reparameterize $\boldsymbol{\mu}_\theta\left(\mathbf{y}_k, k\right)$ as:
\begin{equation}
\mu_\theta\left(\mathbf{y}_k, k\right)=\frac{1}{\sqrt{\alpha_k}}\left(\mathbf{y}_k-\frac{\beta_k}{\sqrt{1-\bar{\alpha}_k}} \epsilon_\theta\left(\mathbf{y}_k, k \right)\right)
\end{equation}
where $y_k=\sqrt{\bar{\alpha}_k} \mathbf{y}_0+\left(1-\bar{\alpha}_k\right) \epsilon$, then the eventual loss is:
\begin{equation}
\begin{aligned}
L(\theta) 
&=\mathbb{E}_q\left[\lambda\left\|\tilde{\boldsymbol{\mu}}_k\left(\mathbf{y}_k, \mathbf{y}_0\right)-\boldsymbol{\mu}_\theta\left(\mathbf{y}_k, k\right)\right\|^2\right] \\
&\propto \mathbb{E}_q\left\|\epsilon-\epsilon_\theta\left(\mathbf{y}_k, k \right)\right\|^2
\end{aligned}
\end{equation}

During inference, we recover original sample $y_0$ from $y_K\sim\mathcal{N}\left(0,\mathbf{I}\right) $ step by step according to:
\begin{equation}
y_{k-1}=\frac{1}{\sqrt{\alpha_k}}\left(\mathbf{y}_k-\frac{\beta_k}{\sqrt{1-\bar{\alpha}_k}} \epsilon_\theta\left(\mathbf{y}_k, k \right)\right)
\end{equation}

\subsection{Conditional Denoising Diffusion Probabilistic Model}
When using the diffusion model for trajectory prediction, it is important to consider the impacts of historical trajectory and environmental information on the future trajectory distribution. Mathematically, we represent all inputs by a context vector $\mathbf{x}$ and aim to analyze the conditional distribution $p_\theta\left( \mathbf{y}_0\mid  \mathbf{x}\right)$. The forwarding process is defined as follows:
\begin{equation}
\label{eq12}
\begin{aligned}
q\left(\mathbf{y}_{1: K}\mid \mathbf{y}_0,\mathbf{x}\right)&:=\prod_{k=1}^K 
q\left(\mathbf{y}_{k} \mid \mathbf{y}_{k-1}\right) \\
q\left(\mathbf{y}_{k} \mid \mathbf{y}_{k-1}\right)&:=\mathcal{N}\left(\mathbf{y}_{k} ;  \sqrt{1-\beta_t} \mathbf{y}_{k-1}, \beta_k \mathbf{I}\right)
\end{aligned}
\end{equation}
while the reverse process is parameterized by:
\begin{equation}
\label{eq13}
\begin{aligned}
p_\theta\left(\mathbf{y}_{0: K}\mid \mathbf{x}\right)&:=p\left(\mathbf{y}_K\mid \mathbf{x}\right) \prod_{k=1}^K 
p_\theta\left(\mathbf{y}_{k-1} \mid \mathbf{y}_k,\mathbf{x}\right) \\
\quad p_\theta\left(\mathbf{y}_{k-1} \mid \mathbf{y}_k,\mathbf{x}\right)&:=\mathcal{N}\left(\mathbf{y}_{k-1} ; \boldsymbol{\mu}_\theta\left(\mathbf{y}_k, \mathbf{x},k\right), \boldsymbol{\Sigma}_\theta\left(\mathbf{y}_k, \mathbf{x},k\right)\right)
\end{aligned}
\end{equation}
Similarly, variational inference is used to maximize $\log p_\theta\left( \mathbf{y}_0 \mid  \mathbf{x} \right)$, and the evidence lower bound is computed as:
\begin{equation}
\label{eq14}
\begin{aligned}
% \mathbb{E}_q\left[\log \frac{p_\theta\left(\mathbf{y}_{0: K}\mid \mathbf{x}\right)}{q\left(\mathbf{y}_{1: K} \mid \mathbf{y}_0, \mathbf{x}\right)}\right] 
ELBO
=\mathbb{E}_q\log p\left(\mathbf{y}_K\mid \mathbf{x}\right)  
+\mathbb{E}_q\sum_{k=1}^K \log \frac{p_\theta\left(\mathbf{y}_{k-1} \mid \mathbf{y}_k,\mathbf{x}\right)}{q\left(\mathbf{y}_k \mid \mathbf{y}_{k-1}\right)}] 
\end{aligned}
\end{equation}

If we choose a large step ${K}$, then the first term can also be ignored like DDPM because:
\begin{equation}
\log p\left(\mathbf{y}_K\mid \mathbf{x}\right)=\log \mathbb{E}_{p\left( \mathbf{y}_0 \mid \mathbf{x}\right)}q\left( \mathbf{y}_K \mid \mathbf{y}_0\right)
\end{equation}
where $q\left( \mathbf{y}_K \mid \mathbf{y}_0\right)$ is always a normal distribution. This is what \cite{gu2022stochastic} does. The final loss function will be:
\begin{equation}
L(\theta) 
= \mathbb{E}_q\left\|\epsilon-\epsilon_\theta\left(\mathbf{y}_k, x, k \right)\right\|^2
\end{equation}
where the noised added in the $k$-th step will be predicted under the guidance of the context vector.

To implement the aforementioned process, it is crucial to ensure that $q\left( \mathbf{y}_K \mid \mathbf{y}_0\right)$ is always a normal distribution under any context vector. This requirement usually necessitates a large number of diffusion steps, which can be inefficient and time-consuming for inference. Additionally, when dealing with complex original distributions, determining the appropriate number of steps for completing the diffusion process is difficult. Therefore, in the following section, we present our solution which aims to generate multimodal trajectory distributions more efficiently.

% \begin{equation}
% \begin{aligned}
% &=\mathbb{E}_{q}\log p_{\boldsymbol{\theta}}\left(y_0 \mid y_1, x\right)-D_{\mathrm{KL}}\left(q\left(y_K \mid y_0,x\right) \| p\left(y_K \mid x\right)\right) \\
% &-\sum \mathbb{E}_{q}\left[D_{\mathrm{KL}}\left(q\left(y_{k-1} \mid y_k, y_0\right) \| p_{\boldsymbol{\theta}}\left(y_{k-1} \mid y_k,x\right)\right)\right]
% \end{aligned}
% \end{equation}

\section{Model description} \label{sec:model}

% In this section, we devise a AutoEncoder-based model to simultaneously detect node-level and edge-level events. 
This section starts with a concise overview of IDM’s architecture. We subsequently delve into more comprehensive explanations of the diffusion processes for goals and trajectories. Finally, we present implementation details.
% Details are expatiated as follows.  

\begin{figure*}[t]
\centering
\includegraphics[width=15cm]{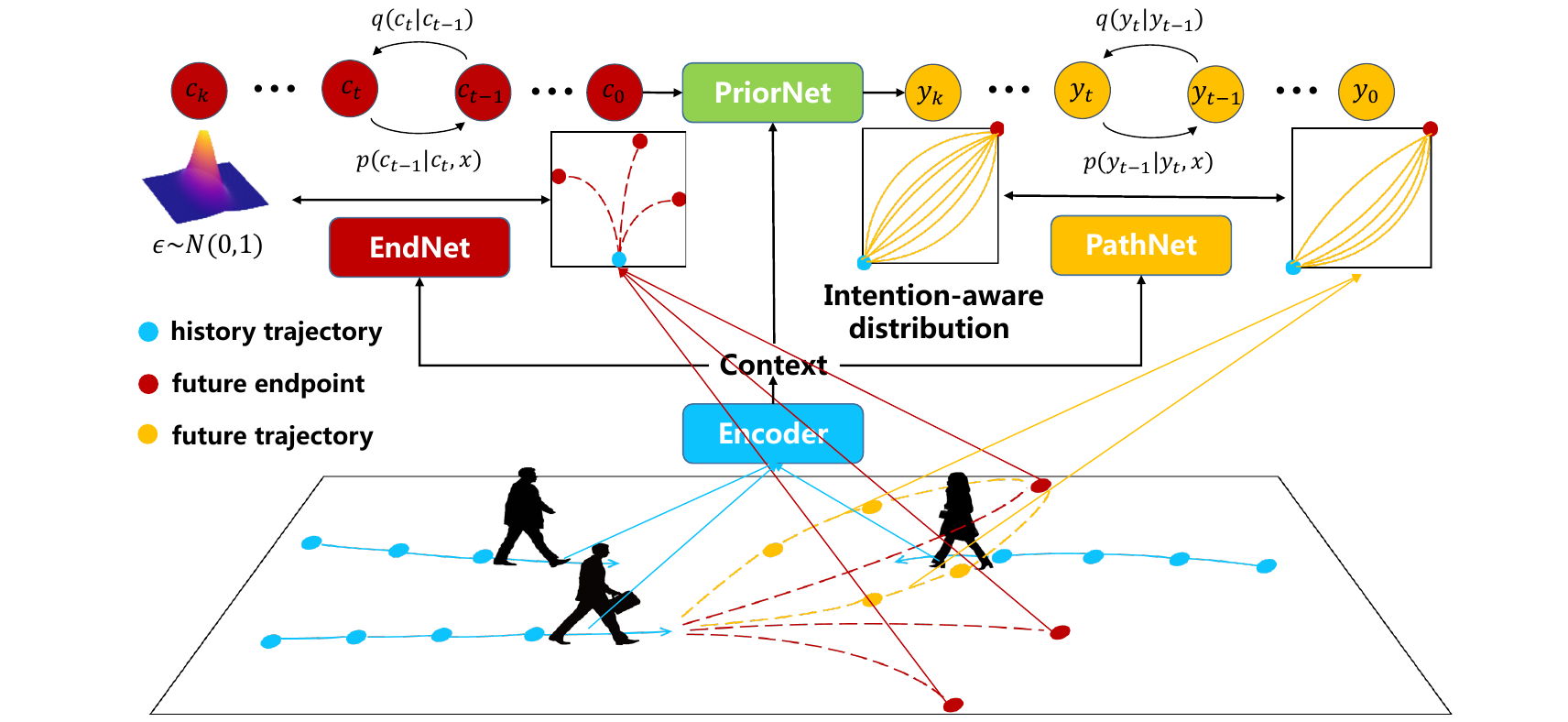}
\caption{Overall structure of intention-aware trajectory diffusion model: a) EndNet: It models the diffusion process of the agent's endpoints. b) PathNet: It models the diffusion process of the agent's trajectories conditioned on a specific endpoint. c) PriorNet: It estimates the initial noise distribution of the trajectory diffusion process with fewer steps.}
\label{fig:model}
\end{figure*}

\subsection{Architecture}
The overall architecture of IDM is presented in Fig~\ref{fig:model}. We follow an encoder-decoder architecture. The encoder takes all observations as inputs and produces a context vector, referred to as the “context” for simplicity. The decoder then transforms this context into multiple future trajectories. In this paper, we focus on designing the decoder, making the model independent of the specific encoder used. The decoder consists of two diffusion processes:
1) Goal Diffusion Process: This process aims to model the uncertainty associated with the goal (endpoint). The denoising function in the reverse process is parameterized by an EndNet. 
2) Trajectory Diffusion Process: This process models the uncertainty of action given a specific goal. To expedite the diffusion process, we develop a PriorNet that estimates the prior noise distribution. Furthermore, we discover that the goal itself can provide clues for the estimation of the prior noise distribution. With the prior noise distribution established, we employ a PathNet to iteratively eliminate noise and generate the final trajectories.

\subsection{Goal Diffusion Process}
The goal-based diffusion process aims to learn the distribution of endpoints under a specific context. This process is denoted as $ \left(c_0,c_1,...,c_{K} \right)$, where $K$ represents the number of diffusion steps, and $c_0$ represents the ground truth distribution. As the diffusion progresses, $c_0$ gradually transitions towards a random noise distribution. The forward process is formulated as follows:
\begin{equation}
\begin{aligned}
q\left(\mathbf{c}_{1: K}\mid \mathbf{c}_0,\mathbf{x}\right)&:=\prod_{k=1}^K 
q\left(\mathbf{c}_{k} \mid \mathbf{c}_{k-1}\right) \\
q\left(\mathbf{c}_{k} \mid \mathbf{c}_{k-1}\right)&:=\mathcal{N}\left(\mathbf{c}_{k} ;  \sqrt{1-\beta_t} \mathbf{c}_{k-1}, \beta_k \mathbf{I}\right)
\end{aligned}
\end{equation}

In our approach, we adopt a large value of $K$, similar to conventional DDPM. As the diffusion process progresses, $c_{K}$ can be considered as following a normal distribution.
According to Sec.~\ref{sec:pre}, the noise injected at each diffusion step is parameterized by a neural network called EndNet, represented as$\epsilon_\theta\left(\mathbf{c}_k, x, k \right)$. To generate the eventual endpoints, we follow a step-by-step procedure, utilizing the EndNet-generated noise. More specifically, the endpoints are generated iteratively as follows:
 \begin{equation}
\mathbf{c}_{k-1}=\frac{1}{\sqrt{\alpha_k}}\left(\mathbf{c}_k-\frac{\beta_k}{\sqrt{1-\bar{\alpha}_k}} \epsilon_\theta\left(\mathbf{c}_k, \mathbf{x}, k \right)\right)
\end{equation}
where $\theta$ represents the parameters of the EndNet. The loss function for the goal diffusion process is:
\begin{equation}\label{eq19}
L_{goal}
= \mathbb{E}_q\left\|\epsilon-\epsilon_\theta\left(\mathbf{c}_k, \mathbf{x}, k \right)\right\|^2
\end{equation}

In this process, the sample being modeled is a 2-D location, which leads to a significant reduction in inference time.
% The efficiency of our model is further detailed in the following parts. 

\subsection{Trajectory Diffusion Process}\label{ssec:tdp}
When the goal is determined, the trajectory of the agent may include uncertainties. For example, a pedestrian may react to the presence of surrounding objects by opting for a moderate or aggressive action to avoid a collision, as illustrated in Fig.~\ref{fig:model}. To address these uncertainties, we introduce an additional diffusion process that is specifically designed to incorporate the agent’s intention. This integration of intention into the diffusion process enables more precise predictions of the agent’s future movements.
% A interesting finding is that intention can also cut back the steps of diffusion process. 

First, the forward process is defined as:
\begin{equation}
\begin{aligned}
q\left(\mathbf{y}_{1: S}\mid \mathbf{y}_0,\mathbf{x}\right)&:=\prod_{k=1}^K 
q\left(\mathbf{y}_{s} \mid \mathbf{y}_{s-1}\right) \\
q\left(\mathbf{y}_{s} \mid \mathbf{y}_{s-1}\right)&:=\mathcal{N}\left(\mathbf{y}_{s} ;  \sqrt{1-\beta_t} \mathbf{y}_{s-1}, \beta_s \mathbf{I}\right)
\end{aligned}
\end{equation}
where $S$ is the number of steps and differs from that of goal diffusion $K$. If $S$ is small, the first term of Eq~\ref{eq14} can not be ignored, because $y_S$ is not a constant normal distribution. The new evidence lower bound is computed as follows:
\begin{equation}
\label{eq22}
\begin{aligned} 
& ELBO \\
&=\mathbb{E}_q\log p\left(\mathbf{y}_S\mid \mathbf{x}\right)  
+\mathbb{E}_q\sum_{s=1}^S \log \frac{p_\theta\left(\mathbf{y}_{s-1} \mid \mathbf{y}_s,\mathbf{x}\right)}{q\left(\mathbf{y}_s \mid \mathbf{y}_{s-1}\right)}\\
&=\mathbb{E}_q\left[\log \frac{p\left(\mathbf{y}_S\mid \mathbf{x}\right)p_\theta\left(\mathbf{y}_0\mid \mathbf{y}_1,\mathbf{x} \right)}{q\left(\mathbf{y}_1 \mid \mathbf{y}_{0}\right)} + \sum_{s=2}^S \frac{p_\theta\left(\mathbf{y}_{s-1}\mid \mathbf{y}_s,x\right)}{q\left(\mathbf{y}_{s}\mid \mathbf{y}_{s-1}\right)} \right]\\
&=\mathbb{E}_q\left[\log \frac{p\left(\mathbf{y}_S\mid \mathbf{x}\right)p_\theta\left(\mathbf{y}_0\mid \mathbf{y}_1,\mathbf{x} \right)}{q\left(\mathbf{y}_S \mid \mathbf{y}_{0}\right)} + \sum_{s=2}^S \frac{p_\theta\left(\mathbf{y}_{s-1}\mid \mathbf{y}_s, x\right)}{q\left(\mathbf{y}_{s-1}\mid \mathbf{y}_{s}, \mathbf{y}_{0}\right)} \right]\\
&=\mathbb{E}_q\log p_\theta\left(\mathbf{y}_0\mid \mathbf{y}_1,\mathbf{x} \right) - 
D_{K L}\left(q\left(\mathbf{y}_{S} \mid \mathbf{y}_0\right)\| p \left(\mathbf{y}_{S} \mid x\right)\right)\\
&- \sum_{s=2}^{S}\mathbb{E}_q\left[ D_{K L}\left(q\left(\mathbf{y}_{s-1} \mid \mathbf{y}_{s},\mathbf{y}_0\right)\| p_\theta \left(\mathbf{y}_{s-1} \mid \mathbf{y}_{s}, x\right)\right) \right]
\end{aligned}
\end{equation}
where the sum of the first and third terms equals the EVLB of the original diffusion process. The second term varies only with $p \left(\mathbf{y}_{S} \mid x\right)$, and a large $S$ leads to an extreme condition where the second term is a constant. Then the loss function of the trajectory diffusion process is defined according to:
\begin{equation}
L_{traj}= L_{diff}+L_{prior}
\end{equation}
where $ L_{diff}$ is the negative sum of the first and third term, and $L_{prior}$ is the negative of the second term. The computation of $L_{diff}$ is the same as in the conventional diffusion process:
\begin{equation}\label{eq23}
L_{diff}
= \mathbb{E}_q\left\|\epsilon-\epsilon_\varphi\left(\mathbf{y}_s, \mathbf{x}, s \right)\right\|^2
\end{equation}
where $\varphi$ denotes the parameters of the network that estimates the noise at each step. 

To compute $L_{prior}$, an intuitive practice is to parameterize the $p \left(\mathbf{y}_{S} \mid x\right)$ with another neural network represented by $\phi$, like what the variational autoencoder does. Considering that $q \left(\mathbf{y}_{S} \mid \mathbf{y}_{0}\right)$ is a g distribution $q\left(\mathbf{y}_S \mid \mathbf{y}_0\right)=\mathcal{N}\left(\mathbf{y}_S ; \sqrt{\bar{\alpha}_S} \mathbf{y}_0,\left(1-\bar{\alpha}_S\right) \mathbf{I}\right)$, according to Eq.~\ref{eq3}, we also take $p \left(\mathbf{y}_{S} \mid x\right)$ as a G distribution. 
The neural network estimates the mean $\mu_{\phi}\left( x \right)$ and variance $\sigma_{\phi}\left( x \right)$ of $p \left(\mathbf{y}_{S} \mid x \right)$. To simplify the computation, we make $\sigma_{\phi}\left( x \right)=\left(1-\bar{\alpha}_S\right) \mathbf{I}$, and compute $L_{prior}$ as follows:
\begin{equation}
L_{prior}=\left\|\mu_{\phi}\left( \mathbf{x} \right)- \sqrt{\bar{\alpha}_S} \mathbf{y}_0\right\|^2
\end{equation}

In this computation, there is a question about whether a Gaussian distribution is sufficient to capture all the information in $p \left(\mathbf{y}_{S} \mid x\right)$. Previous study \cite{zhao2021tnt} highlights that the trajectory distribution can be decomposed into multiple independent distributions conditioned on the goals of the agent.
\begin{equation}
p \left(\mathbf{y}_{0} \mid x\right)= \int p\left(\mathbf{c} \mid x\right)p\left(\mathbf{y}_{0} \mid x,\mathbf{c}\right)d\mathbf{c}
\end{equation}
where $\mathbf{c}$ denotes the goal of the agent. Then $p \left(\mathbf{y}_{S} \mid x\right)$ is formulated as:
\begin{equation}
\begin{aligned}
p \left(\mathbf{y}_{S} \mid x\right)
&= \int p\left(\mathbf{y}_{0} \mid x\right)p\left(\mathbf{y}_{S} \mid x,\mathbf{y}_{0}\right)d\mathbf{y}_{0}\\ 
&= \iint p\left(\mathbf{c} \mid x\right)p\left(\mathbf{y}_{0} \mid x,\mathbf{c}\right)q\left(\mathbf{y}_{S} \mid \mathbf{y}_{0}\right)d\mathbf{c}d\mathbf{y}_{0}\\
&=\int p\left(\mathbf{c} \mid x\right)p\left(\mathbf{y}_{S} \mid x,\mathbf{c}\right)d\mathbf{c}
\end{aligned}
\end{equation}

If we decompose the distribution $p \left(\mathbf{y}_{S} \mid x\right)$ by conditioning on targets, we can calculate $L_{prior}$ as follows:
\begin{equation}
\begin{aligned}
L_{prior}
&=D_{K L}\left(q\left(\mathbf{y}_{S} \mid \mathbf{y}_0\right)\| p \left(\mathbf{y}_{S} \mid x\right)\right)\\
&=D_{K L}\left(q\left(\mathbf{y}_{S} \mid \mathbf{y}_0\right)\| \int p\left(\mathbf{c} \mid x\right)p\left(\mathbf{y}_{S} \mid x,\mathbf{c}\right)d\mathbf{c}\right) \\
&=-\mathbb{E}_{q\left(\mathbf{y}_{S} \mid \mathbf{y}_0\right)} \log \frac{\mathbb{E}_{p\left(\mathbf{c} \mid \mathbf{x}\right)}p\left(\mathbf{y}_{S} \mid x,\mathbf{c}\right)}{q\left(\mathbf{y}_{S} \mid \mathbf{y}_0\right)}\\
&\leq -\mathbb{E}_{q\left(\mathbf{y}_{S} \mid \mathbf{y}_0\right)}\mathbb{E}_{p\left(\mathbf{c} \mid \mathbf{x}\right)}\log\frac{p\left(\mathbf{y}_{S} \mid x,\mathbf{c}\right)}{q\left(\mathbf{y}_{S} \mid \mathbf{y}_0\right)}\\
&= \mathbb{E}_{p\left(\mathbf{c} \mid \mathbf{x}\right)} D_{K L}\left(q\left(\mathbf{y}_{S} \mid \mathbf{y}_0\right)\| p \left(\mathbf{y}_{S} \mid x,\mathbf{c}\right)\right)
\end{aligned}
\end{equation}
where the upper bound of $L_{prior}$ is induced according to Jensen's inequality. During training, we minimize the upper bound of $L_{prior}$, which we call $L_{prior}$ latter.
Instead of modeling the whole trajectory distribution, we parameterize the distribution of the trajectory conditioned on the goal with a neural network $\phi$. The final $L_{prior}$ is calculated by:
\begin{equation}\label{eq:28}
L_{prior} =  \mathbb{E}_{p\left(\mathbf{c} \mid \mathbf{x}\right)}\left\|\mu_{\phi}\left( \mathbf{x}, c \right)- \sqrt{\bar{\alpha}_S} \mathbf{y}_0\right\|^2
\end{equation}

Compared with $p\left(\mathbf{y}_{S} \mid x\right)$, $p\left(\mathbf{y}_{S} \mid x,c\right)$ is more suitable to be modeled as a Gaussian distribution, and could lead to small prior loss. The insight is in accord with previous works\cite{gu2021densetnt,bae2022non}, which regard the future trajectory as a multimodal distribution. Our experiments verify this point in Sec.~\ref{ssec:abla}.  

\subsection{Model details}
A model instantiated from our architecture consists of four blocks: Encoder, EndNet, PriorNet, and PathNet, which we will elaborate literally as follows. 

\subsubsection{Encoder}
Our architecture focuses on the decoder based on the diffusion process and therefore is encoder-agnostic. In this work, we utilize the encoder from a classical model called Trajectron++ \cite{salzmann2020trajectron++}. Trajectron++ follows a CVAE framework. The encoder in Trajetron++ incorporates all observations into a hidden embedding, while the decoder transforms the embedding into plenty of future trajectories, It consists of three main blocks: 1) LSTM network modeling the agent history; 2) Aggregation block that encodes agent interactions; 3) CNN based map encoder which represents the semantic maps by a vector. The final output of the encoder is concatenated by the above embeddings. We follow their encoder settings.

\subsubsection{EndNet}
EndNet is responsible for parameterizing $\epsilon_\theta\left(\mathbf{c}_k, x, k \right)$. It takes $x$, step $k$, and the goal at that step $\mathbf{c}_k$ as inputs, and produces the estimated noise. These inputs are concatenated and fed into a Multi-Layer Perceptron (MLP). We evaluate the performance of EndNet with different numbers of layers in Sec.~\ref{ssec:abla}.  

\subsubsection{PriorNet}
PriorNet is used to model $\mu_{\phi}\left( x, \mathbf{c} \right)$, which represents the mean of the prior distribution $p\left(\mathbf{y}_{S} \mid x\right)$. The inputs include the context vector $x \in \mathbb{R}^{D} $ and the goal of the agent $\mathbf{c} \in \mathbb{R}^{2}$. The output of PriorNet is $\mathbf{y}_{S} \in \mathbb{R}^{T_Q \times2}$, where $T_{Q}$ represents the number of future time steps. There could exist temporal correlations among future trajectories. Therefore, we generate the prior distribution based on a neural network that captures temporal correlation. MLP, RNN, TCN, and Transformer are all validated in our experiments. 

% Additionally, inspired by \cite{lim2021temporal}, we devise a gated network to capture the correlation. Lim et.al proposed a gated 
% residual network to automatically determine the extent of required non-linear processing, where the simple linear correlation can be modeled by a residual connection. In trajectory prediction, given a fixed goal, the trajectory is most likely a straight line without any interference. However, the trajectory will be non-linear when considering the environment interaction like collision avoidance. Hence, we devise a gate network to handle both linear and non-linear correlation.

\subsubsection{PathNet}
PathNet tries to parameterize the $\epsilon_\varphi\left(\mathbf{y}_s, x, s \right)$. The network takes $\mathbf{y}_s \in\mathbb{R}^{T_Q\times2} $, $x \in \mathbb{R}^{D}$ and $s$ as inputs and produces the estimated noise $\epsilon \in\mathbb{R}^{T_Q\times2} $. To utilize the temporal correlation, we also develop several versions of PathNet based on MLP, RNN, TCN, and Transformer.

\subsection{Training and Inference}
% In this part, we first introduce the procedures of training and inference and then analyze the computational complexity of our method.  

\subsubsection{Training}
During training, we know both the historical observations and future trajectories. The training loss is defined by:
\begin{equation}
L_{total} = L_{goal} + \lambda_1 L_{diff} + \lambda_2 L_{prior}
\end{equation}
where $L_{goal}$ and $L_{diff}$ are calculated according to Eq.~\ref{eq19} and Eq.~\ref{eq23}. $\lambda_1 $ and $\lambda_2$ control the importance of the loss for the trajectory diffusion process and loss for prior distribution estimation, respectively. 
Because $\mathbf{c}_k$ is sampled from a Gaussian distribution based on $\mathbf{c}_0$ according to Eq.~\ref{eq3}, which is non-differentiable during training, we apply the reparameterization trick. Specifically, we get a sample $\epsilon$ from normal distribution, and obtain $\mathbf{c}_k$ by:
\begin{equation}
\mathbf{c}_{k} = \sqrt{\bar{\alpha}_k} \mathbf{c}_0 +\left(1-\bar{\alpha}_k\right)\epsilon
\end{equation}
$\mathbf{y}_s$ in $L_{diff}$ is acquired by the same way.

Similar to previous works\cite{zhao2021tnt}, we acquire $L_{prior}$ by a teacher-forcing manner, where the realistic goal $\mathbf{c}_0$ is used to guide the calculation:
\begin{equation}\label{eq31}
L_{prior} =  \left\|\mu_{\phi}\left( x, \mathbf{c}_0 \right)- \sqrt{\bar{\alpha}_S} \mathbf{y}_0\right\|^2
\end{equation}
The complete training procedure is shown in Algorithm 1.

\begin{algorithm}[H]
\caption{Training}\label{alg:alg1}
\begin{algorithmic}
\STATE 
% \STATE \textbf{Input}: Dataset $D=\left\{d^{i}=(x^{i},e^{i},\mathcal{M}^{i},y^{i})\mid i=1,2,...B\right\}$
\STATE \textbf{Input}: $x$, $e$, $\mathcal{M}$, $y$; $\alpha_{1:K}^{g}$, $\alpha_{1:S}^{t}$; $\lambda_1$, $\lambda_2$
\STATE \textbf{Output}: Encoder $g_\psi$, EndNet $\epsilon_\theta$, PriorNet $\mu_\phi$, PathNet $\epsilon_\varphi$
\STATE \textbf{repeat}
\STATE \hspace{0.5cm} Compute context vector $\mathbf{x}=g_\psi(x,e,\mathcal{M})$
% \STATE \hspace{0.5cm} $\mathbf{c}_0 \sim q(\mathbf{c}_0)$,  $\mathbf{y}_0 \sim q(\mathbf{y}_0)$
\STATE \hspace{0.5cm} $\mathbf{y}_0=y, \mathbf{c}_0=y[-1]$
\STATE \hspace{0.5cm} $k \sim$ Uniform$(\left\{1,2,...,K\right\})$,  $s \sim$ Uniform$(\left\{1,2,...,S\right\})$
\STATE \hspace{0.5cm} $\epsilon \sim \mathcal{N}(\mathbf{0},\mathbf{I})$
\STATE \hspace{0.5cm} $\mathbf{c}_k=\sqrt{\bar{\alpha}_k^{g}} \mathbf{c}_0  +\left(1-\bar{\alpha}_k^{g}\right) \epsilon$
\STATE \hspace{0.5cm} $\mathbf{y}_s=\sqrt{\bar{\alpha}_s^{t}} \mathbf{y}_0  +\left(1-\bar{\alpha}_s^{t}\right) \epsilon$
\STATE \hspace{0.5cm} Compute $L_{goal},L_{diff}, L_{prior}$ according to Eq.~\ref{eq19},\ref{eq23},\ref{eq31}
% \STATE \hspace{0.5cm} $L_{goal} =\left\|\epsilon-\epsilon_\theta\left(\sqrt{\bar{\alpha}_k} \mathbf{c}_0 \epsilon +\left(1-\bar{\alpha}_k\right) \mathbf{I}, x, k \right)\right\|^2$
\STATE \hspace{0.5cm} Take gradient descent step on 
\STATE \hspace{1.0cm} $\nabla_{\psi,\theta,\varphi,\phi}(L_{goal}+\lambda_1 L_{diff}+ \lambda_2 L_{prior})$
\STATE \textbf{until} converged
\end{algorithmic}
\label{alg1}
\end{algorithm}

\subsubsection{Inference}
In the inference phase, we aim to generate $\mathcal{K}$ plausible predictions. The goal is predicted by the iterative diffusion process first, and is then utilized to initialize $\mathbf{y}_S$. Finally, the trajectory conditioned on the goal is generated by another diffusion process with fewer steps. The details are described in Algorithm 2.

\begin{algorithm}[H]
\caption{Inference}\label{alg:alg2}
\begin{algorithmic}
\STATE 
\STATE \textbf{Input}: $x$, $e$, $\mathcal{M}$, $y$; $\alpha_{1:K}^{g}$, $\alpha_{1:S}^{t}$, $g_\psi$, $\epsilon_\theta$, $\mu_\phi$, $\epsilon_\varphi$
\STATE \textbf{Output}: $\left\{y_{i} \in \mathbb{R}^{T_{Q}\times2} \mid i=1,2,...\mathcal{K}\right\}$
\STATE Compute context vector $\mathbf{x}=g_\psi(x,e,\mathcal{M})$
\STATE \textbf{for} $i = 1,...,\mathcal{K}$ \textbf{do}
\STATE \hspace{0.5cm} $\mathbf{c}_K \sim \mathcal{N}(\mathbf{0},\mathbf{I})$
\STATE \hspace{0.5cm} \textbf{for} $k = K,...,1$ \textbf{do}
\STATE \hspace{1.0cm} $\mathbf{c}_{k-1}=\frac{1}{\sqrt{\alpha_k^{g}}}\left(\mathbf{c}_k-\frac{\beta_k^{g}}{\sqrt{1-\bar{\alpha}_k^{g}}} \epsilon_\theta\left(\mathbf{c}_k, \mathbf{x}, k \right)\right)$
\STATE \hspace{0.5cm} \textbf{end for}
\STATE \hspace{0.5cm} $\epsilon \sim \mathcal{N}(\mathbf{0},\mathbf{I})$
\STATE \hspace{0.5cm} $\mathbf{y}_S=\mu_\phi(\mathbf{x},\mathbf{c}_0)+ \left(1-\bar{\alpha}_S\right)\epsilon$
\STATE \hspace{0.5cm} \textbf{for} $s = S,...,1$ \textbf{do}
\STATE \hspace{1.0cm} $\mathbf{y}_{s-1}=\frac{1}{\sqrt{\alpha_s^{t}}}\left(\mathbf{y}_s-\frac{\beta_s^{t}}{\sqrt{1-\bar{\alpha}_s^{t}}} \epsilon_\varphi\left(\mathbf{y}_s, \mathbf{x}, s \right)\right)$
\STATE \hspace{0.5cm} \textbf{end for}
\STATE \hspace{0.5cm} $y_i=\mathbf{y}_0$
\STATE \textbf{end for} 
\STATE \textbf{return} $\left\{y_{i} \in \mathbb{R}^{T_{Q}\times2} \mid i=1,2,...\mathcal{K}\right\}$ 
% \STATE \hspace{0.5cm}$ N_\mathbf{t} \gets | \{ i : \mathbf{t}_i = \mathbf{t} \} | $ \textbf{ for } $ \mathbf{t}= -1,+1 $
% \STATE \hspace{0.5cm}$ B_i \gets \sqrt{ \textsc{max}(N_{-1},N_{+1}) / N_{\mathbf{t}_i} } $ \textbf{ for } $ i = 1,...,N $
% \STATE \hspace{0.5cm}$ \hat{\mathbf{H}} \gets  B \cdot (\mathbf{X}^T\textbf{W})/( \mathbb{1}\mathbf{X} + \mathbb{1}\textbf{W} - \mathbf{X}^T\textbf{W} ) $
% \STATE \hspace{0.5cm}$ \beta \gets \left ( I/C + \hat{\mathbf{H}}^T\hat{\mathbf{H}} \right )^{-1}(\hat{\mathbf{H}}^T B\cdot \mathbf{T})  $
% \STATE \hspace{0.5cm}\textbf{return}  $\textbf{W},  \beta $
% \STATE 
% \STATE {\textsc{PREDICT}}$(\mathbf{X} )$
% \STATE \hspace{0.5cm}$ \mathbf{H} \gets  (\mathbf{X}^T\textbf{W} )/( \mathbb{1}\mathbf{X}  + \mathbb{1}\textbf{W}- \mathbf{X}^T\textbf{W}  ) $
% \STATE \hspace{0.5cm}\textbf{return}  $\textsc{sign}( \mathbf{H} \beta )$
\end{algorithmic}
\label{alg2}
\end{algorithm}

\section{Experiments}

\subsection{Experiment Setup}

\subsubsection{Dataset}
We use Stanford Drone Datasets (SDD) and UCY/ETH to evaluate our methods.

\textbf{SDD}: It is a large-scale dataset aimed at trajectory prediction. The data is collected by a drone from a bird’s-eye view perspective. It contains 60 recordings referring to 20 scenes around a university campus and describes the movements of multiple pedestrians and vehicles. The trajectory data is sampled at 2.5 FPS. Following previous works\cite{chiara2022goal}, we conduct experiments on 47 recordings about pedestrian trajectory, with 30 of them used as training data, while the remaining 17 recordings are used to test the model’s performance.

\textbf{ETH/UCY}: The dataset is a classical benchmark for pedestrian trajectory prediction. The data comes from the surveillance video on the street. The dataset comprises five scenes, namely, ETH, Hotel, UNIV, ZARA1, and ZARA2. The sampling frequency is 2.5 FPS. We adopt the leave-one-scene-out setting like \cite{gupta2018social}, where 4 scenes are used for training and the remaining scene is used for testing.

\begin{table}[h]
\setlength{\tabcolsep}{3mm}
\caption{Performance on SDD dataset}\label{tab:sdd}
\begin{tabular}{c|ccc|cc}
\hline
Methods                         & Deocder   & \multicolumn{1}{l}{$U_{goal}$} & \multicolumn{1}{l|}{$U_{traj}$} & \multicolumn{1}{l}{ADE} & \multicolumn{1}{l}{FDE} \\ \hline
Social-LSTM                     & BG        & -           & \checkmark          & 57.00             & 31.20                   \\
\multicolumn{1}{l|}{Expert+GMM} & BG        & -           & \checkmark          & 10.67             & 14.38                   \\
SimAug                          & Grid        & \checkmark  & -                   & 10.27             & 19.71                   \\
PCCSNET                         & Grid        & \checkmark  & -                   & 8.62              & 16.16                   \\
Y-Net                           & Grid        & \checkmark  & \checkmark          & 8.97              & 14.61                   \\
Social-GAN                      & GAN       & -           & \checkmark          & 27.23             & 41.44                   \\
Goal-GAN                        & GAN       & \checkmark  & -                   & 12.20             & 22.10                   \\
MG-GAN                          & GAN       & -           & \checkmark          & 13.60             & 25.80                   \\
CGNS                            & CVAE      & -           & \checkmark          & 15.60             & 28.20                   \\
Trajectron++                    & CVAE      & -           & \checkmark          & 8.98              & 19.02                   \\
PECNET                          & CVAE      & \checkmark  & -                   & 9.96              & 15.88                   \\
LB-EBM                          & Energy    & \checkmark  & \checkmark          & 8.87              & 15.61                   \\
MID                             & Diffusion & -           & \checkmark          & 7.61              & 14.30                   \\ \hline
IDM                            & Diffusion & \checkmark  & \checkmark          & \textbf{7.46}      & \textbf{13.83}                   \\ \hline
\end{tabular}
\end{table}

\subsubsection{Metrics}
To evaluate the effectiveness of our model, ADE and FDE are used to measure the accuracy of predictions. Average Displacement Error (ADE) represents the average point-to-point Euclidean distance between the predicted trajectory and the ground truth. Final Displacement Error (FDE) describes the error between the predicted endpoint and the realistic endpoint. The error is measured by pixel in SDD and by meter in ETH/UCY. Given multiple predictions, the best-of-N strategy\cite{gupta2018social} is utilized to compute the final metric. 

\subsubsection{Implementation Details}
During the training stage, the batch size is set to 256, the learning rate is set to 0.0001, $\lambda_1$ is set to 1, and $\lambda_2$ is set to 0.5. All experiments are conducted on a single RTX 3090. During the prediction stage, we generate 20 predictions for each agent and calculate the minimum ADE/FDE based on these predictions. For each dataset, we repeat the evaluation five times and report the average results.

\begin{table*}[h]
\centering
\caption{Performance on ETH/UCY datasets}\label{tab:eth}
\begin{tabular}{c|cc|cccccccccc|cc}
\hline
\multirow{2}{*}{Methods} & \multirow{2}{*}{Deocder} & \multicolumn{1}{l|}{\multirow{2}{*}{Sample}} & \multicolumn{2}{c}{ETH} & \multicolumn{2}{c}{HOTEL} & \multicolumn{2}{c}{UNIV} & \multicolumn{2}{c}{ZARA1} & \multicolumn{2}{c|}{ZARA2} & \multicolumn{2}{c}{AVG} \\ \cline{4-15} 
                         &                          & \multicolumn{1}{l|}{}                        & ADE        & FDE        & ADE         & FDE         & ADE         & FDE        & ADE         & FDE         & ADE          & FDE         & ADE        & FDE        \\ \hline
Social-LSTM  & BG & 20        & 1.09 & 2.35 & 0.79 & 1.76 & 0.67 & 1.40 & 0.47 & 1.00 & 0.56 & 1.17 & 0.72 & 1.54       \\
Social-STGCN & BG & 20        & 0.64 & 1.11 & 0.49 & 0.85 & 0.44 & 0.79 & 0.34 & 0.53 & 0.30 & 0.48 & 0.44 & 0.75       \\
Causal-STGCN & BG & 20        & 0.64 & 1.00 & 0.38 & 0.45 & 0.49 & 0.81 & 0.34 & 0.53 & 0.32 & 0.49 & 0.43 & 0.66       \\
STAR         & BG & 20        & 0.36 & 0.65 & 0.17 & 0.36 & 0.31 & 0.62 & 0.26 & 0.55 & 0.22 & 0.46 & 0.26 & 0.53       \\
Expert-GMM   & BG & 20        & 0.37 & 0.65 & \textbf{0.11} & \textbf{0.15} & 0.20 & 0.44 & 0.15 & 0.31 & 0.12 & 0.26 & \textbf{0.19} & 0.36       \\
PCCSNET      & Grid & 20      & \textbf{0.28} & 0.54 & \textbf{0.11} & 0.19 & 0.29 & 0.60 & 0.21 & 0.44 & 0.15 & 0.34 & 0.21 & 0.42       \\
% Y-net        & Grid & 10000   & 0.28 & 0.33 & 0.10 & 0.14 & 0.24 & 0.41 & 0.17 & 0.27 & 0.13 & 0.22 & 0.18 & 0.27       \\
Social-GAN   & GAN & 20       & 0.81 & 1.52 & 0.72 & 1.61 & 0.60 & 1.26 & 0.34 & 0.69 & 0.42 & 0.84 & 0.58 & 1.18       \\
Goal-GAN     & GAN & 20       & 0.59 & 1.18 & 0.19 & 0.35 & 0.60 & 1.19 & 0.43 & 0.87 & 0.32 & 0.65 & 0.43 & 0.85       \\
Social-BiGAT & GAN & 20       & 0.69 & 1.29 & 0.49 & 1.01 & 0.55 & 1.32 & 0.30 & 0.62 & 0.36 & 0.75 & 0.48 & 1.00       \\
MG-GAN       & GAN & 20       & 0.47 & 0.91 & 0.14 & 0.24 & 0.54 & 1.07 & 0.36 & 0.73 & 0.29 & 0.60 & 0.36 & 0.71       \\
CGNS         & CVAE & 20      & 0.62 & 1.40 & 0.70 & 0.93 & 0.48 & 1.22 & 0.32 & 0.59 & 0.35 & 0.71 & 0.49 & 0.97       \\
PECNET       & CVAE & 20      & 0.54 & 0.87 & 0.18 & 0.24 & 0.35 & 0.60 & 0.22 & 0.39 & 0.17 & 0.30 & 0.29 & 0.48       \\
Trajectron++ & CVAE & 20      & 0.39 & 0.83 & 0.12 & 0.21 & \textbf{0.20} & 0.44 & \textbf{0.15} & 0.33 & \textbf{0.11} & \textbf{0.25} & \textbf{0.19} & 0.41       \\
LB-EBM       & Energy & 20    & 0.30 & \textbf{0.52} & 0.13 & 0.20 & 0.27 & 0.52 & 0.20 & 0.37 & 0.15 & 0.29 & 0.21 & 0.38       \\
MID          & Diffusion & 20 & 0.39 & 0.66 & 0.13 & 0.22 & 0.22 & 0.45 & 0.17 & 0.30 & 0.13 & 0.27 & 0.21 & 0.38\\ \hline
IDM         & Diffusion & 20 & 0.41 & 0.62 & 0.15 & 0.25 & \textbf{0.20} & \textbf{0.42} & 0.17 & \textbf{0.28} & 0.12 & \textbf{0.25} &  0.21 & \textbf{0.36} \\ \hline
\end{tabular}
\end{table*}

\subsection{Comparison with SOTA methods}

The results for the SDD dataset are presented in Table~\ref{tab:sdd}. The table includes all the methods along with their decoder type and whether they model the uncertainty of the goal and trajectory. The Bivariate Gaussian (BG) models assume that future positions follow a bivariate Gaussian distribution and utilize maximum likelihood estimation to compute the parameters of the distribution. The methods make strong assumptions about the trajectory distribution, which can result in a poor fit to real data. The grid-based models divide the scene into grid cells and predict the occupancy probability of each cell in the future. While these methods achieve outstanding performance, they require significant effort to construct and train the trajectory map \cite{mangalam2021goals}. Generative models like GAN and CVAE are also used to generate future trajectories. However, training GAN-based models can be uncontrollable \cite{guo2022end}, while the CVAE-based models tend to generate unnatural trajectories \cite{gu2022stochastic}. Overall, our method achieves the best performance among all existing methods in both metrics, with an ADE of 7.46 and an FDE of 13.83 in pixel coordinates. In addition, two findings can be acquired from the results. First, MID and IDM perform better than all other kinds of methods, demonstrating that diffusion models have great potential in trajectory prediction. Second, our model also outperforms MID. This is because 
MID does not consider the uncertainty of the goal, and can therefore lose accuracy, especially in long-term prediction. Our model reduces the FDE by 0.47 compared to MID. Hence, it is essential to model the uncertainty of agents' intentions and trajectories.
% Third, the grid-based methods second only to diffusion model and outperform other generation based model (i.e. CVAE based, GAN based model). 

Table~\ref{tab:eth} presents the results on ETH/UCY dataset. The prediction results of IDM are comparable to those of existing methods, with an average ADE of 0.21 and FDE of 0.36. While IDM may not improve the MID in all datasets, it achieves similar performance with less time and memory requirement, as demonstrated in Sec.~\ref{sec:eff}. Furthermore, compared to MID, our method can generate more natural trajectories, as shown in Sec.~\ref{sec:vis}.

\subsection{Visualization}\label{sec:vis}
\begin{figure*}[t]
\centering
\includegraphics[width=18cm]{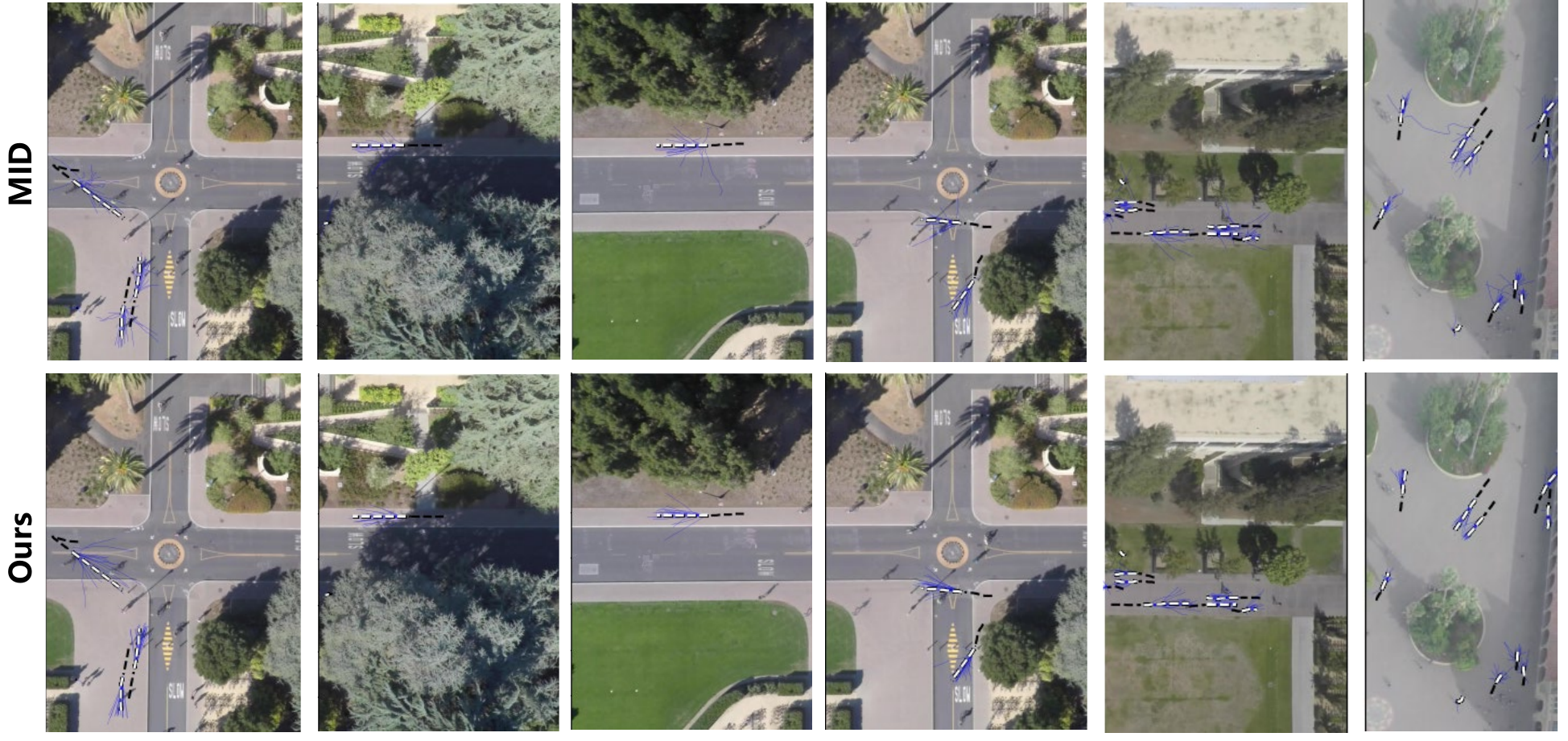}
\caption{Prediction on SDD}
\label{fig:vis_sdd}
\end{figure*}

\begin{figure*}[t]
\centering
\includegraphics[width=18cm]{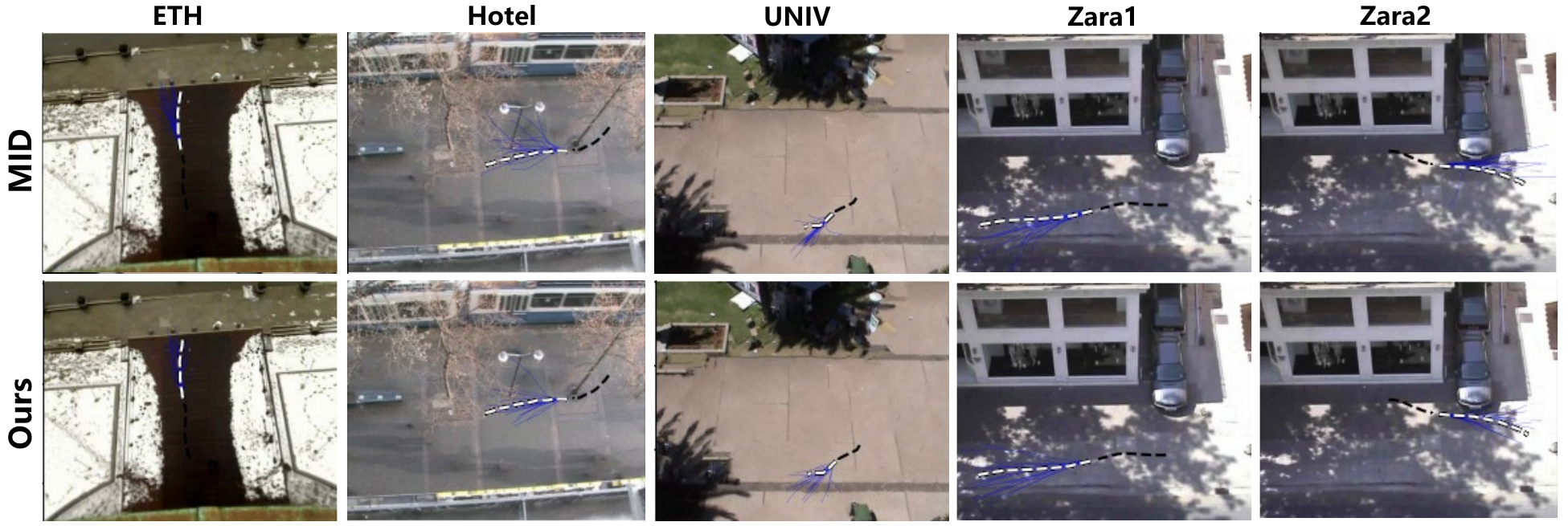}
\caption{Prediction on ETH/UCY}
\label{fig:vis_eth}
\end{figure*}

In order to investigate the diversity and accuracy of our model, we generate 20 predictions using both our model and MID on all datasets and visualize the predictions. Fig.~\ref{fig:vis_sdd} and Fig.~\ref{fig:vis_eth} display the results on SDD and ETH/UCY, respectively. The black dashed line represents the historical trajectory, the white dashed line represents the ground truth future trajectory, and the blue solid line represents the predicted trajectory. Both MID and IDM generate diverse predictions, including some that closely approach the ground truth future trajectories. However, MID tends to produce some unnatural trajectories. For example, on SDD datasets, a pedestrian is predicted to turn around in the first case and abruptly steer in the third case. Additionally, the predictions generated by MID sometimes appear uneven as seen in the sixth case where a pedestrian is expected to follow a twisted trajectory. Similar results have been observed in predictions on ETH/UCY datasets. In contrast,  IDM tends to generate smoother and more realistic trajectories that resemble plausible future movements.

To quantify these differences, we reduce the number of predictions and observe the minimum ADE and FDE of the two methods. The results are presented in Table~\ref{tab:nat}. Our method consistently outperformed MID in both metrics, regardless of the number of predictions made. Furthermore, the superiority of our method becomes more distinct as the number of samples decreases. This is because MID is more prone to predicting unnatural trajectories that deviate significantly from the ground truth. When making fewer predictions, there is a higher likelihood those correct and natural predictions may be excluded. Our model mitigates this problem by incorporating the predicted goals into the generation of trajectories.

\begin{figure*}[t]
\centering
\includegraphics[width=18cm]{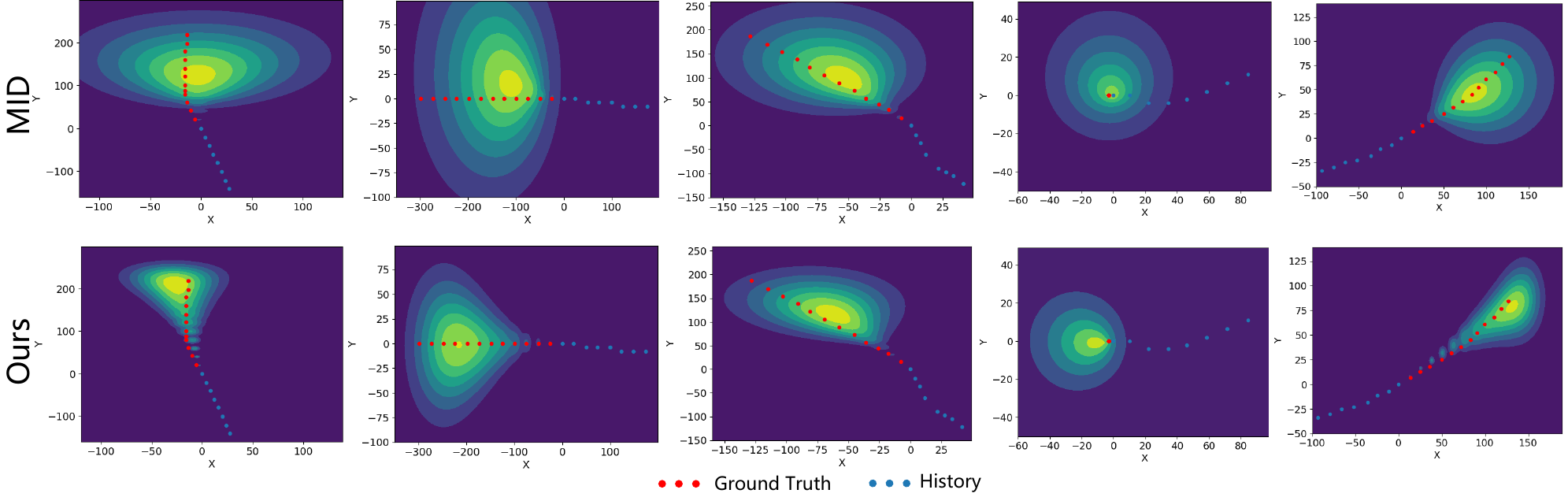}
\caption{Counter map of predictions}
\label{fig:counter}
\end{figure*}

We also present the predicted trajectory distribution of MID and our method as contours in Fig.~\ref{fig:counter}. It can be observed that, compared with MID, our method predicts a smaller walkable region, indicating that more implausible trajectories can be excluded. Furthermore, the uncertainty of the trajectory predicted by our method increases with the prediction step, which aligns with our intuition. In contrast, the predictions of MID present high uncertainty even though in the early future. Additionally, as shown in the fourth case, MID may generate unnatural distributions where pedestrians have a higher chance of turning around. Overall, by introducing intention information, our method enables predictions that better fit the real distribution.

\begin{table}[h]
\caption{ADE/FDE under different number of predictions}\label{tab:nat}
\scalebox{0.93}{
\begin{tabular}{cccccc}
\hline
\multicolumn{1}{l}{} & \multicolumn{1}{c}{4} & \multicolumn{1}{c}{8} & \multicolumn{1}{c}{12} & \multicolumn{1}{c}{16} & \multicolumn{1}{c}{20} \\ \hline
MID                  & 12.40/27.45           & 10.11/21.30           & 9.09/19.27            & 8.48/16.42             & 7.61/14.30             \\
Ours                 & 11.19/24.58           & 9.44/19.74           & 8.38/17.79             & 7.85/15.80             & 7.46/13.83             \\ \hline
\end{tabular}
}
\end{table}

\subsection{Ablation Study}\label{ssec:abla}
In this section, we begin by analyzing the role of each component in our architecture. Next, we validate the selection of each component. Finally, we evaluate the impact of the step number on the prediction results.

\begin{figure*}[h]
\centering
\subfigure[PriorNet]{
\includegraphics[width=6cm]{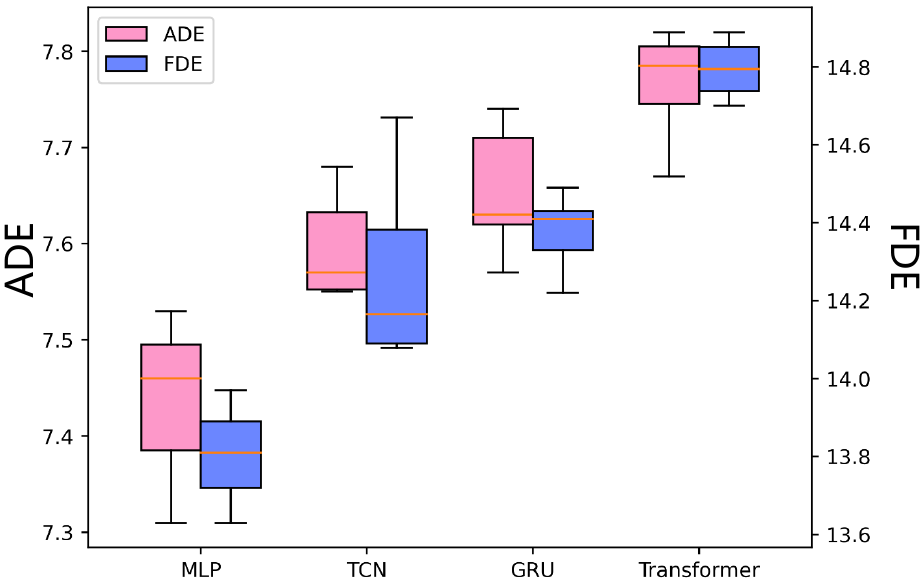}
%\caption{fig1}
\label{fig:abla1}
}%
\subfigure[PathNet]{
\includegraphics[width=6cm]{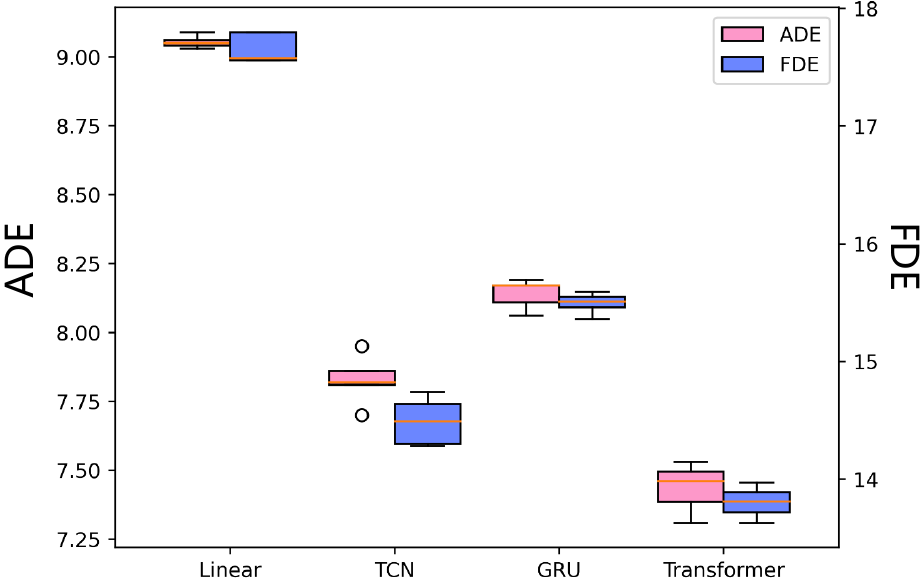}
%\caption{fig2}
\label{fig:abla2}
}%
\subfigure[EndNet]{
\includegraphics[width=6cm]{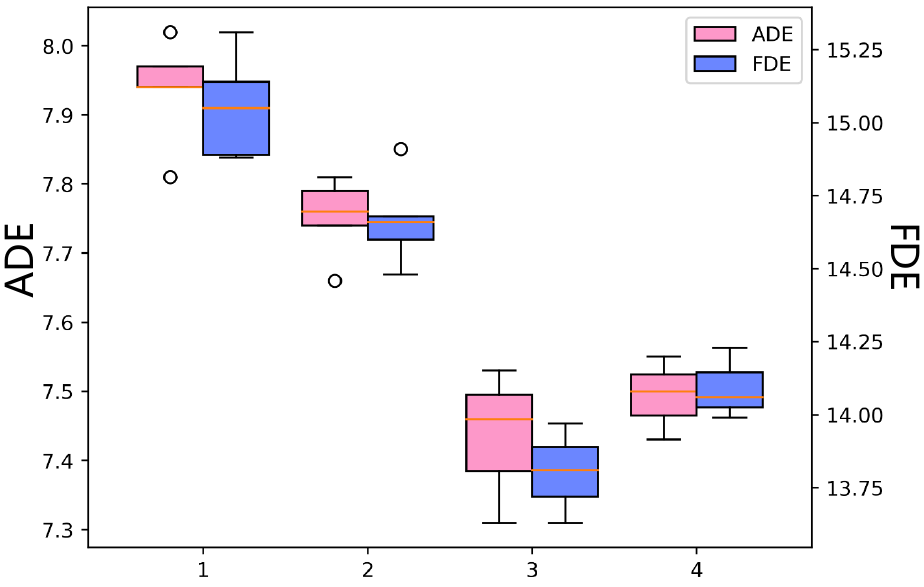}
%\caption{fig2}
\label{fig:abla3}
}%
\caption{Selection of each component}
\end{figure*}

Table~\ref{tab:abla} presents the prediction results when each component is absent. First, the variants without the PriorNet apply fewer steps in the diffusion process and still consider initial signals as Gaussian noise. During the training process, it does not consider the prior loss in Eq.~\ref{eq:28}. This variant exhibits the highest prediction error in both ADE and FDE. This highlights the importance of estimating the initial distribution for the diffusion process with fewer steps. The second variant only models the distribution of the goal and uses MLP to generate trajectories conditioned on the given goal. The variant achieves a higher ADE and FDE because it ignores the action uncertainty, which can be modeled by another diffusion process. The third variant models the action uncertainty with the trajectory diffusion process. However, the estimation of the prior distribution solely relies on the context vector and does not take into account any information about the goal. Our model surpasses it by 6.8\% in ADE and 10.3\% in FDE. This demonstrates that incorporating goal information can provide crucial cues for accurately estimating the prior distribution of the diffusion process.

To make the most of our model's efficacy, we explore different selections for each component. Fig.~\ref{fig:abla1} shows the performance of different PriorNet. Surprisingly, the MLP selection achieves the best performance. This suggests that, given a fixed goal, the most probable prior trajectory distribution is simple, and complex networks are more susceptible to overfitting. For PathNet, we try similar networks, but find that the complex Transformer performs better than other choices (Fig.~\ref{fig:abla2}). It indicates that Transformer excels at modeling the uncertainty of trajectories that exhibit temporal correlation. We also experiment with different layers in the MLP for EndNet and find that three is the most suitable number (Fig.~\ref{fig:abla3}). Increasing the number of layers does not result in improved performance.

\begin{figure*}[h]
\centering
\subfigure[Efficiency comparison for different datasets]{
\includegraphics[width=6cm]{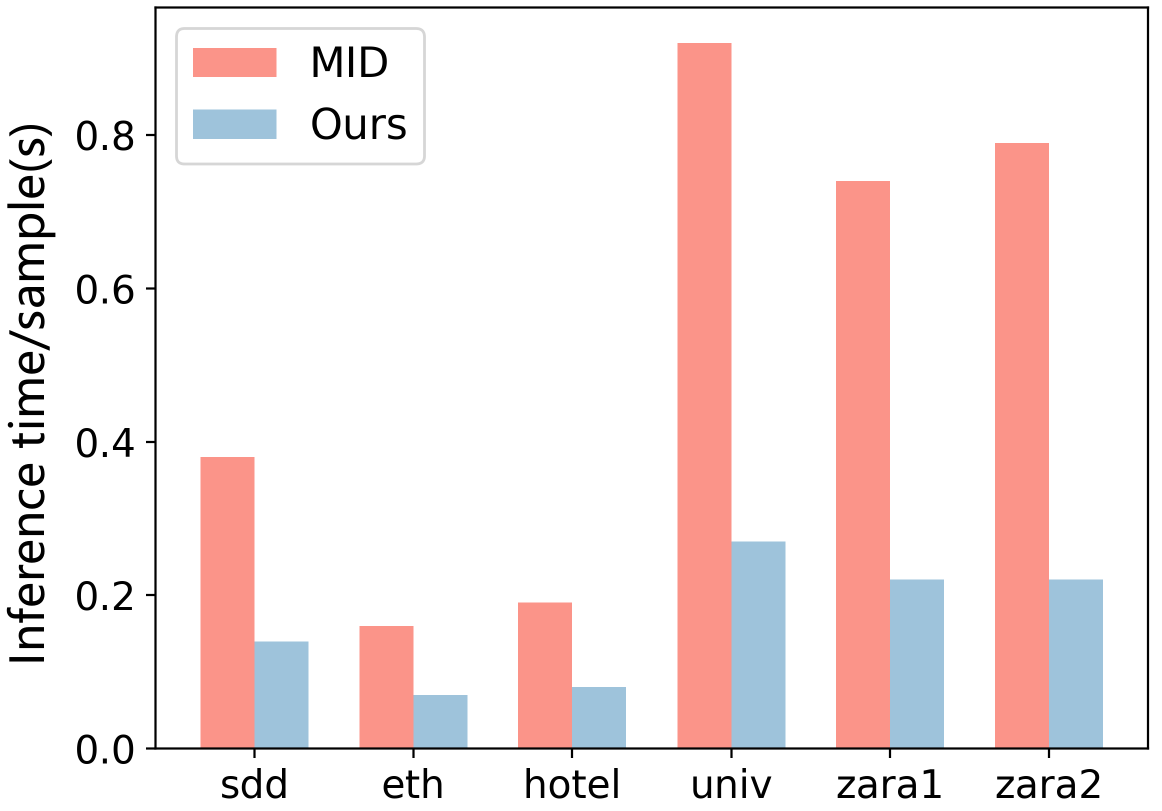}
%\caption{fig1}
\label{fig:vel1}
}%
\subfigure[Efficiency under different steps $\tau$]{
\includegraphics[width=6cm]{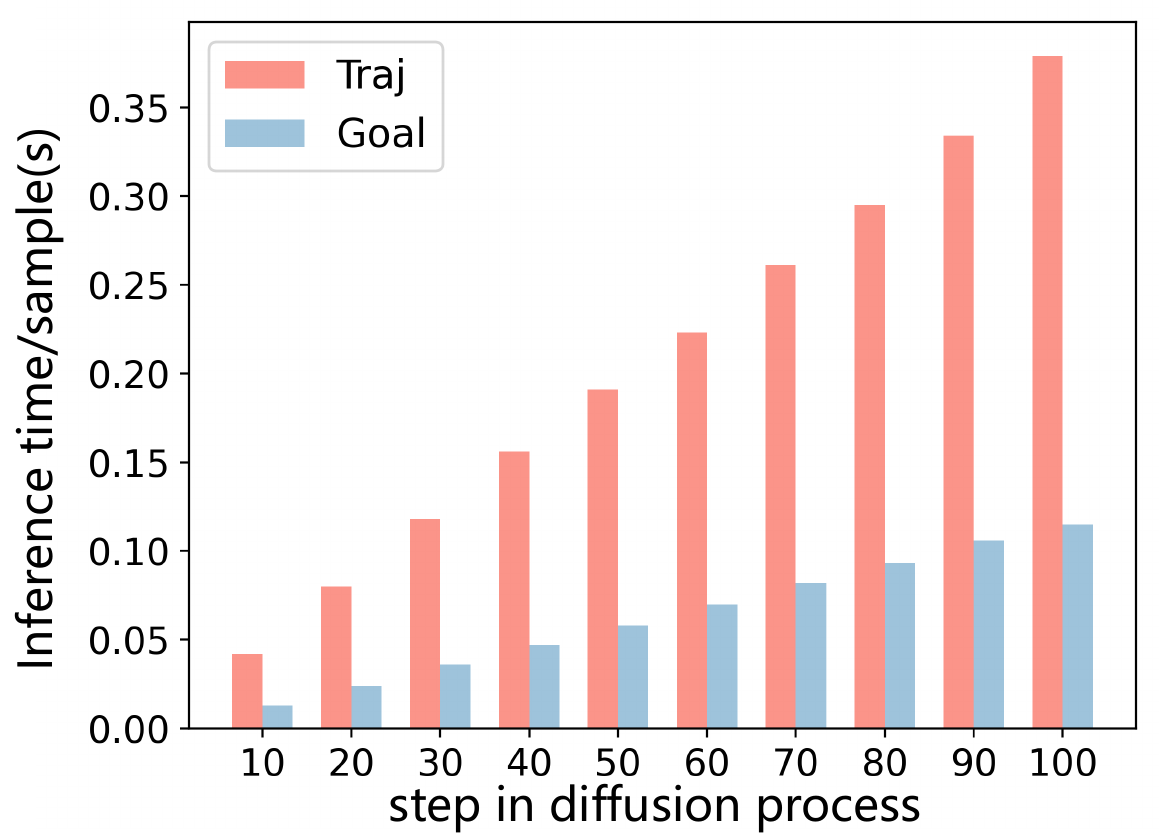}
%\caption{fig2}
\label{fig:vel2}
}%
\subfigure[Total performance under different $\tau$]{
\includegraphics[width=5.5cm]{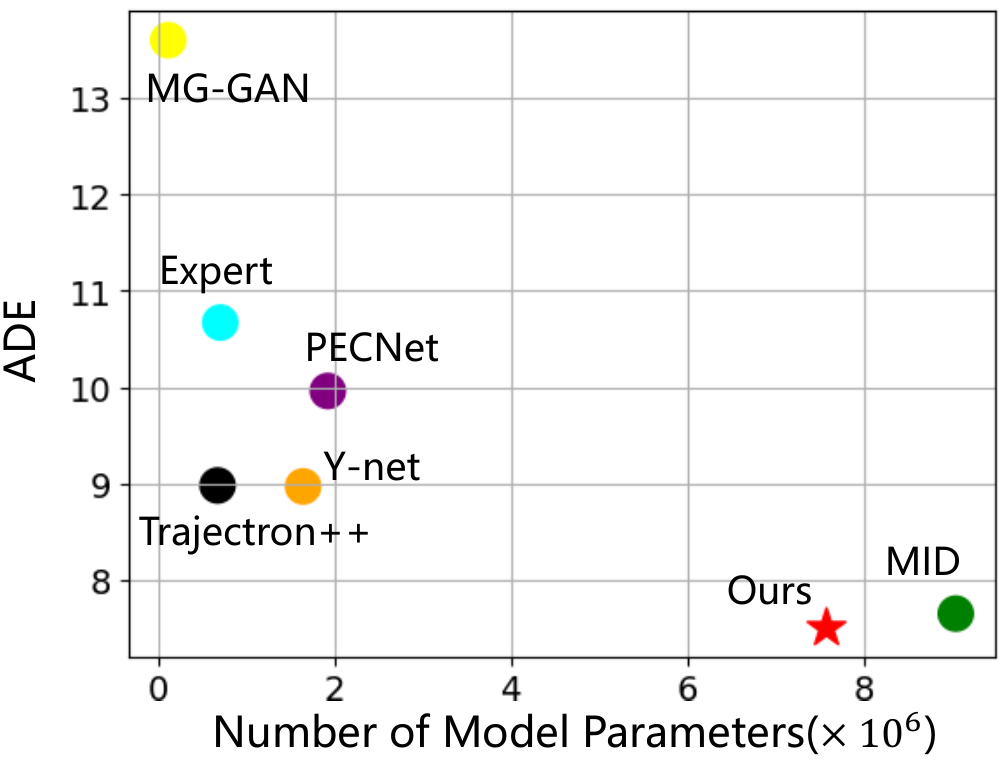}
%\caption{fig2}
\label{fig:params}
}%
\caption{Inference time and model parameter}
\end{figure*}

\begin{figure}[h]
\centering
\includegraphics[width=8cm]{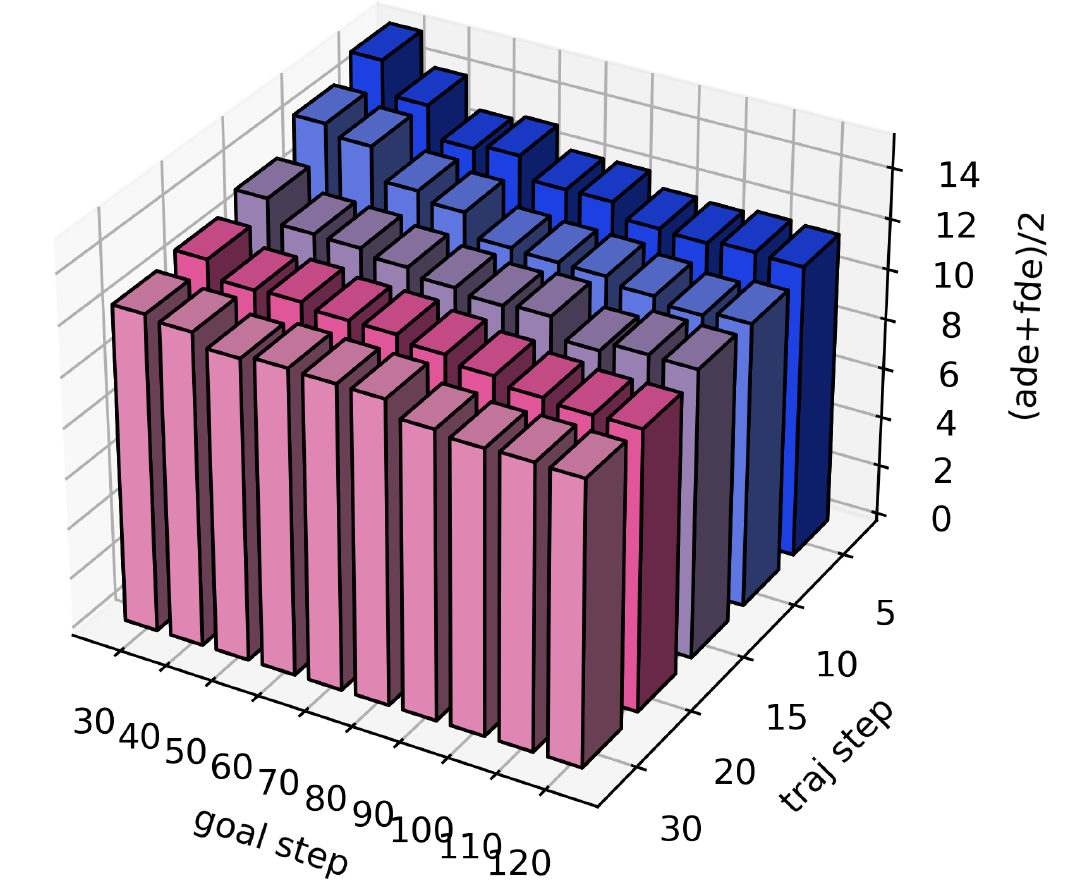}
%\caption{fig1}
\centering
\caption{Selection of diffusion steps for goal and trajectory}
\label{fig:step}
\end{figure}

Furthermore, we discuss the influence of step numbers on the model. Fig.~\ref{fig:step} shows the performance under different steps of the goal and trajectory diffusion process. Firstly, increasing the number of steps in the goal diffusion process results in more precise predictions, because accurate goal prediction helps in estimating the prior distribution in the trajectory diffusion process. Secondly, given the specific goal, the prediction result is not susceptible to the number of steps in the trajectory diffusion process. This is because our PriorNet can predict the suitable prior distribution for different steps. The best result, as reported above, is achieved when the goal and trajectory steps are set to 100 and 10, respectively.  

\begin{table}[h]
\caption{Performance influenced by each component}\label{tab:abla}
\setlength{\tabcolsep}{4.6mm}
\begin{tabular}{ccccc}
\hline
\multicolumn{1}{c}{EndNet} & \multicolumn{1}{c}{PriorNet} & \multicolumn{1}{c}{PathNet} & \multicolumn{1}{c}{ADE} & \multicolumn{1}{c}{FDE} \\ \hline
$\checkmark$   &  $-$   &  $\checkmark$   & 9.40 & 17.34    \\
$\checkmark$   &  $-$   &  $-$  & 7.85 & 15.44             \\
${-}$    &  $\checkmark$ &  $\checkmark$   & 7.98 & 15.41  \\
$\checkmark$   & \checkmark & $\checkmark$   & 7.46  & 13.83    \\ \hline
\end{tabular}
\end{table}

\subsection{Efficiency}\label{sec:eff}

We compare the inference speed of our model with that of the conventional diffusion model. Fig.~\ref{fig:vel1} demonstrates the execution time per prediction of the two methods on different datasets. Compared to MID, our method experiences nearly a threefold decrease in inference time. This superiority stems from two reasons. Firstly, our goal diffusion process handles 2-D locations, whereas MID handles sequences. Secondly, our model completes the trajectory diffusion process in just one-tenth of the steps. Additionally, in the scenes with high population density, the inference time further increases. Across all datasets, our method achieves a prediction in less than 240ms. Fig.~\ref{fig:vel2} illustrates the inference time of our model with different steps of the goal and trajectory diffusion process. As the number of steps increases, the inference time demonstrates an approximately linear increase. Moreover, when comparing the same number of steps, goal diffusion requires significantly less time than the trajectory diffusion process, indicating that reducing the target dimension is effective in shortening the time.

Furthermore, we also summarize the model parameters of our model. As shown in Fig.~\ref{fig:params}, our model reduces the parameters by one-fifth compared to the original MID. However, the superiority of the diffusion model in trajectory prediction comes at the expense of more model parameters. In the future, we plan to propose the development of a lightweight diffusion model for trajectory prediction.

% \begin{figure}[h]
% \centering
% \includegraphics[width=8cm]{pictures/vel1.pdf}
% %\caption{fig1}
% \centering
% \caption{Efficiency comparison between MID and our method $\tau$}
% \label{fig:vel1}
% \end{figure}

% \begin{figure}[h]
% \centering
% \includegraphics[width=8cm]{pictures/vel2.pdf}
% %\caption{fig1}
% \centering
% \caption{Efficiency under different steps for diffusion process $\tau$}
% \label{fig:vel2}
% \end{figure}

% Here, we provide a complexity analysis mathematically.

% We also compute the model parameters of our methods. Compared to the original MID, our method drops the parameters by one five. However, the superiority of the diffusion model on trajectory prediction comes with a cost of higher model parameters. In the future, we will propose to develop a lightweight diffusion model for trajectory prediction utilizing knowledge distillation.
% \begin{figure}[h]
% \centering
% \includegraphics[width=8cm]{pictures/params.pdf}
% %\caption{fig1}
% \centering
% \caption{Total performance under different $\tau$}
% \label{fig:params}
% \end{figure}

% \usepackage{multirow}

% \begin{figure*}[h]
% \centering
% \subfigure[Total recall]{
% \centering
% \includegraphics[width=9cm]{pictures/exp3.pdf}
% %\caption{fig1}
% \label{fig:exp3}
% }%
% \subfigure[Total delay]{
% \centering
% \includegraphics[width=9cm]{pictures/exp4.pdf}
% %\caption{fig1}
% \label{fig:exp4}
% }%
% \centering
% \caption{Model comparison results on Metr-LA}
% \label{fig:com2}
% \end{figure*}

\section{Conclusion}
In this work, we design an intention-aware diffusion model for multimodal trajectory prediction. The uncertainty of trajectories is decoupled into goal uncertainty and action uncertainty, and is modeled by two interconnected diffusion processes. To improve the efficiency of the inference process, instead of assuming a normal prior noise distribution, we devise a PriorNet to estimate
the specific prior distribution of the diffusion process, thus reducing the required number of steps for a complete diffusion process.  Additionally, we augment the original loss function by incorporating the estimation error of the prior distribution. We conduct experiments on two real-world datasets, SDD and ETH/UCY. Our method achieves state-of-the-art results. Compared to the original diffusion model, our approach reduces the inference time by two-thirds. Furthermore, our experiments reveal the benefits of introducing intention information in modeling the diffusion process with fewer steps.

% \section*{Acknowledgments}
% This should be a simple paragraph before the References to thank those individuals and institutions who have supported your work on this article.

\bibliographystyle{ieeetr} %unsrt
\bibliography{idm}

\vfill

\end{document}